\pdftrailerid{}
\documentclass{article}

\PassOptionsToPackage{numbers, compress}{natbib}
\usepackage[main, final]{neurips_2026}

\usepackage[utf8]{inputenc} \usepackage[T1]{fontenc}    \usepackage{hyperref}       \usepackage{url}            \usepackage{booktabs}       \usepackage{amsfonts}       \usepackage{microtype}      \PassOptionsToPackage{table,dvipsnames}{xcolor}
\usepackage{xcolor}         
\usepackage{graphicx}

\usepackage{amsmath}
\usepackage{amssymb}

\usepackage{tikz}
\DeclareRobustCommand{\gfcircle}{\tikz[baseline=-0.6ex]{\fill[gray] (0,0) circle (0.65ex);}}
\DeclareRobustCommand{\gfcross}{\tikz[baseline=-0.6ex]{\draw[gray,line width=0.45ex]
    (-0.7ex,-0.7ex)--(0.7ex,0.7ex)
    (-0.7ex,0.7ex)--(0.7ex,-0.7ex);}}
\DeclareRobustCommand{\gftriangle}{\tikz[baseline=-0.6ex]{\fill[gray]
    (0,0.75ex) -- (-0.75ex,-0.55ex) -- (0.75ex,-0.55ex) -- cycle;}}

\newcommand{\yrcite}[1]{\citeyearpar{#1}}

\usepackage{wrapfig}        \usepackage{needspace}      \usepackage{adjustbox}
       \usepackage{multirow}
\usepackage{makecell}       

\usepackage{array}
\usepackage{enumitem}
\usepackage{pifont} \usepackage{caption}

\usepackage{threeparttable}

\newcommand{\boldres}[1]{{\textbf{\textcolor{red}{#1}}}}
\newcommand{\secondres}[1]{{\underline{\textcolor{blue}{#1}}}}

\hypersetup{hidelinks,
colorlinks=true,
linkcolor=red,
urlcolor= RoyalBlue,citecolor=cyan
}

\title{MSAR: Next-Scale Autoregressive Forecasting for Time Series via Modular Multi-Scale Decoupling}

\author{
\small 
Fanda Fan$^{1}$\thanks{Equal contributions.} \quad Kuiye Ding$^{1}$\footnotemark[1] \quad Liming Mao$^{2}$ \quad Bingsong Wang$^{2}$ \quad Yao Wang$^{1}$ \quad Xiaorui Wang$^{1, 5}$\\
\small
\textbf{Ruijie Jian$^{5}$ \quad Zhipeng Liu$^{4}$ \quad Luqi Gong$^{2, 3}$\thanks{Corresponding author.} \quad Zhenghua Lu$^{1}$ \quad Chunjie Luo$^{1}$ \quad Jianfeng Zhan$^{1, 5}$}
\\[2pt] 
\small
$^{1}$ Institute of Computing Technology, Chinese Academy of Sciences
\\[2pt]
\small
$^{2}$ Beijing University of Posts and Telecommunications
\hspace{2em} $^{3}$ Zhejiang Lab
\\[2pt] 
\small
$^{4}$ Northeastern University
\hspace{2em} $^{5}$ University of Chinese Academy of Sciences
\\[2pt] 
\small 
\{fanfanda, luochunjie, zhanjianfeng\}@ict.ac.cn 
\hspace{2em} luqi@bupt.edu.cn \\
}

\begin{document}

\maketitle

\begin{abstract}
  Time series forecasting underpins critical applications in finance, energy, healthcare, and transportation. Although deep models have achieved strong results, most adopt single-scale modeling or restrict multiscale processing to the input side, causing a misalignment between multiscale inputs and single-scale outputs and limiting predictive power. We introduce the \textbf{Modular Scale-wise Autoregressive Framework (MSAR)}, a model-agnostic design that forecasts progressively across multiple temporal resolutions. MSAR offers three advantages: (1) \textbf{scale-wise aligned modeling}, which disentangles heterogeneous temporal patterns by aligning inputs and outputs at each scale; (2) \textbf{scale-wise autoregression}, where coarse-scale predictions guide finer-scale forecasting through hierarchical information flow; and (3) a \textbf{modular architecture}, enabling seamless integration with diverse backbones such as CNNs, MLPs, and Transformers. Extensive experiments across a broad set of datasets and forecasting models demonstrate that MSAR achieves consistent improvements in both accuracy and inference efficiency, validating the effectiveness of scale-aligned autoregression for multiscale time series forecasting. The code is available at: \url{https://github.com/Oryxo/MSAR}.
\end{abstract}

\section{Introduction}

Time series forecasting is critical in domains such as finance, energy, healthcare, and transportation. Recent deep models—including RNNs~\cite{salinas2020deepar,bergsma2022cfar,bergsma2023sutranets,lin2023segrnn}, CNNs~\cite{wu2022timesnet,wang2023micn}, MLPs~\cite{zhang2022less,zeng2023transformers, yu2024revitalizing,tang2025unlocking}, and Transformers~\cite{nie2022time,liu2024itransformer,brigato2025position}—have shown strong performance. However, they typically rely on single-scale modeling, processing historical data at a fixed temporal scale. This overlooks a key property of real-world time series: temporal patterns vary across scales~\cite{wang2024timemixer,liu2026timeformer}. Forcing heterogeneous patterns into a unified representation often causes scale interference, leading to poor generalization and degraded forecasting accuracy.

To address the limitations of single-scale modeling, prior works~\cite{geva1998scalenet,guo2023multivariate,zhang2023crossformer,wang2024timemixer, shabani2023scaleformeriterativemultiscalerefining, chen2024pathformermultiscaletransformersadaptive,wang2024timemixer2,han2025retrieval,chen2025simpletm} introduced multiscale-mixing in input modeling to capture patterns at different scales. However, they still predict future values at a single (fine) scale, leading to a mismatch between multiscale inputs and single-scale outputs. Similarly, multi-resolution decoders~\cite{challu2023nhits,kraus2025xlstmmixer} generate coarse-to-fine intermediate components, but these are ultimately \emph{interpolated} into a single-resolution prediction with no per-scale supervision. This misalignment complicates the learning process and weakens the model’s ability to leverage multiscale information effectively, often resulting in suboptimal forecasts.

\begin{figure*}[t]
\begin{center}
\includegraphics[width=1.0\linewidth,height=0.3928877292\linewidth]{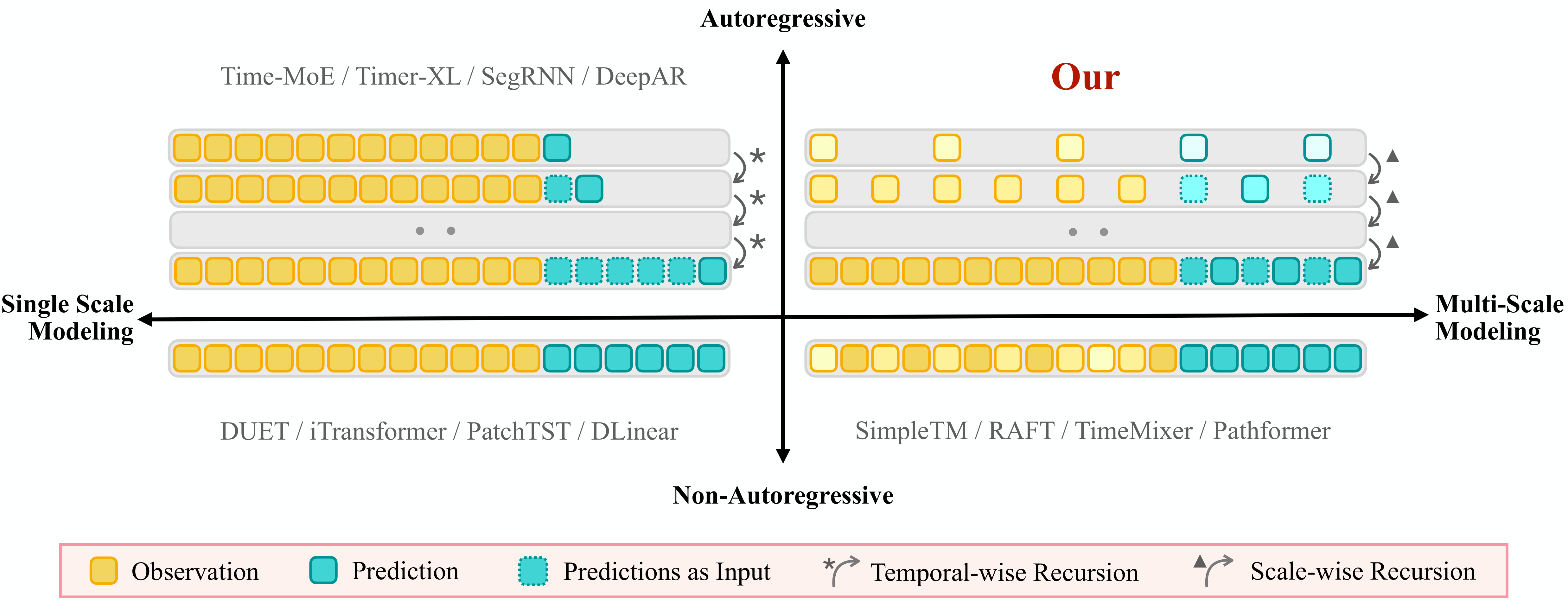}
\end{center}
\caption{Comparison of forecasting paradigms. MSAR introduces scale-wise autoregression with aligned multi-scale inputs and outputs. Coarse-scale predictions serve as contextual guidance for finer-scale forecasting through hierarchical information flow.}
\label{fig:concept-map}
\end{figure*}

Building upon the aforementioned limitations, we propose the Modular Scale-wise Autoregressive Framework (MSAR). As shown in Figure~\ref{fig:concept-map}, MSAR performs \textbf{scale-aligned} and \textbf{decoupled modeling}, where each scale is modeled independently. This explicit alignment helps isolate heterogeneous temporal patterns and simplifies the learning process. Motivated by Tian et al.~\yrcite{tian2024visual}, we introduce a \textbf{scale-wise autoregressive} forecasting strategy. 
This structured design clearly distinguishes MSAR from prior approaches that rely on single-scale decoding~\cite{shi2025timemoe,liu2025timerxl,qiu2025duet,liu2024itransformer,nie2022time,zeng2023transformers}, from cross-scale mixing without resolution consistency~\cite{chen2025simpletm,han2025retrieval,wang2024timemixer,chen2024pathformermultiscaletransformersadaptive}, and from multiresolution decoders~\cite{challu2023nhits,kraus2025xlstmmixer}. In contrast, MSAR enforces \textbf{explicit per-scale supervision} and \textbf{strict within-scale alignment}, enabling principled hierarchical autoregressive refinement across scales.

Specifically, MSAR adopts a scale-wise modular framework that enables flexible implementation across different forecasting models. This design makes the MSAR \textbf{model-agnostic} and easy to integrate into a variety of backbone architectures. To coordinate the prediction process across scales, MSAR introduces a simple yet effective information flow mechanism, where predictions from coarser scales serve as contextual signals to guide finer-scale forecasting. Furthermore, to mitigate potential errors propagated from earlier predictions, we incorporate a lightweight refinement module that selectively fuses coarse predictions with current-scale inputs. This refinement process improves prediction accuracy and enhances consistency across scales. 

MSAR offers three key advantages over prior approaches:
\begin{itemize}[leftmargin=*,itemsep=-0.1em]
\item[\ding{182}] Scale-wise aligned modeling: Unlike previous methods that apply multiscale-mixing~\cite{chen2025simpletm,han2025retrieval,wang2024timemixer,chen2024pathformermultiscaletransformersadaptive} only to input features, MSAR performs fully separated modeling at each scale, with inputs and outputs aligned within the same scale. This design reduces learning complexity by isolating heterogeneous temporal patterns and avoiding interference across scales.
\item[\ding{183}] Scale-wise autoregression: MSAR adopts a scale-wise autoregressive strategy~\cite{tian2024visual}, where predictions from coarser scales serve as contextual signals for finer-scale forecasting. This hierarchical structure improves the model’s ability to capture long-term dependencies while enabling progressive refinement of short-term patterns, thereby enhancing both accuracy and interpretability.
\item[\ding{184}] Modular and model-agnostic design: Each scale in MSAR is processed by an independent forecasting module without architectural constraints. This modular design enables seamless integration with diverse backbone models—such as CNNs~\cite{wu2022timesnet}, MLPs~\cite{zeng2023transformers}, and Transformers~\cite{nie2022time,liu2024itransformer}, and allows flexible adaptation to varying input lengths.

\end{itemize}

\section{Related Work}

\subsection{Time Series Forecasting}

\paragraph{Single-Scale Modeling.}
Modern time series forecasting is predominantly driven by deep learning models, including RNN-based~\cite{lin2023segrnn}, CNN-based~\cite{wu2022timesnet}, Transformer-based~\cite{nie2022time,liu2024itransformer}, and MLP-based~\cite{zeng2023transformers,qiu2025duet} architectures. However, these methods typically operate at a single temporal scale and fail to capture distinct patterns that emerge across different resolutions. As a result, long-context inputs often lead to overfitting, as the model attempts to fit both global trends and local fluctuations within a unified representation, without scale-aware separation~\cite{liu2025timerxl}.

\paragraph{Multiscale Modeling.}
To capture temporal dynamics across different frequencies, several recent works have introduced multiscale modeling into time series forecasting.
SimpleTM~\cite{chen2025simpletm} employs wavelet-based tokenization to generate scale-specific representations for Transformers. 
RAFT~\cite{han2025retrieval} leverages retrieval across different periodicities to inject external multi-scale patterns into forecasting. 
TimeMixer~\cite{wang2024timemixer} employs grouped mixing layers to model dependencies across temporal resolutions in a unified encoder. 
Pathformer~\cite{chen2024pathformermultiscaletransformersadaptive} constructs adaptive pathways to route information across temporal resolutions and distances. These methods effectively encode multiscale patterns in the \emph{input sequence} but still predict at a \emph{single temporal scale}, creating a mismatch between input and output resolutions. 
In contrast, our MSAR performs \emph{scale-aligned}, coarse-to-fine \emph{autoregressive} decoding: each stage only consumes inputs at its assigned resolution and is causally conditioned on the previous (coarser) stage, enabling consistent multiscale modeling across both input and output without fusion-based mismatch.

\paragraph{Multi-Resolution Decoders.}
Many recent architectures employ \emph{multi-resolution decoders}, where intermediate temporal scales are generated or refined before producing the final forecast. Crossformer~\cite{zhang2023crossformer} and MR-Transformer~\cite{zhu2023mr} construct hierarchical or segment-level multi-resolution pathways, yet only the \emph{final fine-scale output} is supervised. Scaleformer~\cite{shabani2023scaleformeriterativemultiscalerefining} performs iterative multi-scale decoding, but all intermediate resolutions are fused into a single prediction through pooling or learned interpolation. xLSTM-Mixer~\cite{kraus2025xlstmmixer} mixes multi-span temporal memories inside a multi-resolution decoding stack but still emits a single-resolution forecast. N-HiTS~\cite{challu2023nhits} explicitly reconstructs intermediate blocks at multiple resolutions, but these are subsequently upsampled and aggregated rather than decoded as separate supervised outputs.
MuSiCNet~\cite{liu2024musicnet} introduces a gradual coarse-to-fine framework for irregularly sampled multivariate time series, progressively refining latent representations to handle temporal sparsity.
Thus, although these methods use multi-resolution \emph{decoding mechanisms}, none of them performs \emph{explicit per-scale decoding with per-scale targets}.

\subsection{Autoregressive Approaches for Sequence Modeling}

Autoregressive (AR) modeling has been a widely adopted paradigm in time series forecasting. DeepAR~\cite{salinas2020deepar} and SutraNets~\cite{bergsma2023sutranets} generate the future sequence one time step at a time, where each prediction depends on historical inputs and previously generated values. SegRNN~\cite{lin2023segrnn} extends conventional RNNs with segment-wise recurrence and parallel multi-step decoding. C2FAR~\cite{bergsma2022cfar} adopts a hierarchical binning strategy to generate each value autoregressively from coarse to fine levels of numerical precision. Despite their differences, all these methods operate at a \emph{single temporal scale}, which limits their ability to capture complex temporal dependencies that span across multiple resolutions. Recent time-series foundation models such as Time-MoE~\cite{shi2025timemoe} and Timer-XL~\cite{liu2025timerxl} also adopt autoregression over time steps, leveraging large model capacity and extensive pretraining to achieve strong performance. However, their focus is on scaling parameters and training data, whereas our work centers on paradigm improvements.

Motivated by VAR~\cite{tian2024visual}, we propose the MSAR, which formulates forecasting as a \emph{next-scale prediction} task across scale-aligned temporal resolutions. MSAR takes as input a historical time series and decomposes both the inputs and prediction targets into multiple aligned temporal scales. Each scale is modeled separately using a plug-in forecasting module, and predictions from coarser scales are progressively used to condition finer-scale forecasts, enabling a hierarchical flow of information.

\section{Methodology}

\paragraph{Problem Formulation}
Let the time series data be represented as: $\mathbf{X}_{1:T} = \{ \mathbf{x}_1, \mathbf{x}_2, \dots, \mathbf{x}_T \} \in \mathbb{R}^{T \times C}$, where $T$ denotes the length of the look-back window and $C$ represents the number of input channels. The goal is to predict the future sequence: $\mathbf{Y}_{1:H} \in \mathbb{R}^{H \times C}$, where $H$ is the forecast horizon. For simplicity, we directly use \( \mathbf{X} \) and \( \mathbf{Y} \) to represent the input and output data, respectively.

\paragraph{Notations}
We define $\mathbf{X}^{\Delta}_{1:T} \in \mathbb{R}^{\lfloor\frac{T}{\Delta}\rfloor \times C}$ as the downsampled sequence over time steps, where $\lfloor\frac{T}{\Delta}\rfloor$ represents the dimension after downsampling. Similarly, we define $\mathbf{Y}^{\Delta}_{1:H} \in \mathbb{R}^{\lfloor\frac{H}{\Delta}\rfloor \times C}$ as the prediction results with a time interval of $\Delta$.

\begin{figure*}[t]
\begin{center}
\includegraphics[width=0.95\linewidth,height=0.4122500808\linewidth]{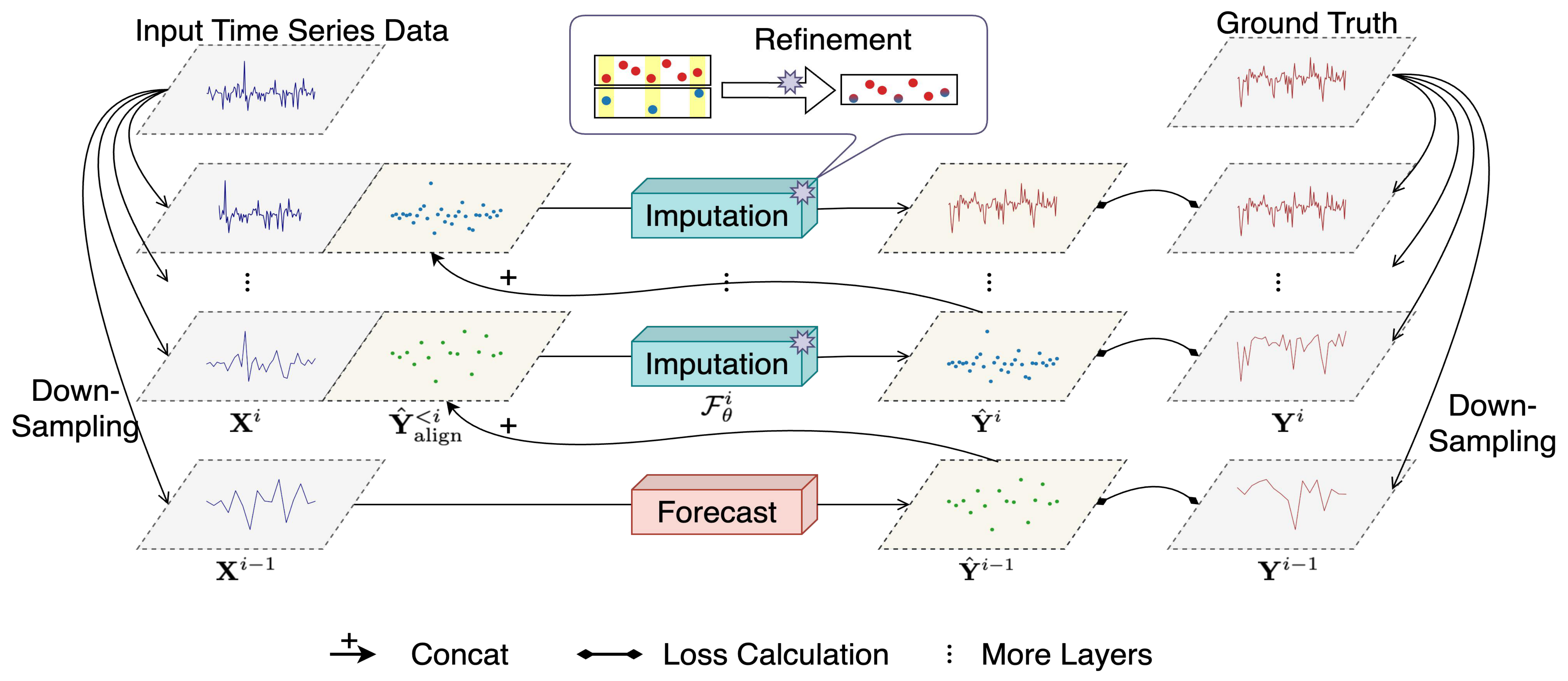}
\end{center}
\caption{
\textbf{Overview of the MSAR framework.}
MSAR is a \textbf{model-agnostic}, \textbf{scale-wise autoregressive pipeline} over multi-resolution inputs obtained by stride down-sampling.
The pipeline starts from the coarsest scale and progressively proceeds to finer scales, where each module \textbf{concatenates} current-scale inputs with \textbf{aligned coarse predictions} and performs \textbf{imputation-based refinement}.
Through this \textbf{progressive imputation} process, MSAR constructs scale-aligned inputs and produces \textbf{coarse-to-fine multi-scale forecasts}.
}
\label{fig:main-paradigm}
\end{figure*}

\subsection{Technical Motivation}
\label{sec:method1}
\begin{wraptable}{r}{0.5\textwidth}
\vspace{-25pt}
\centering
\caption{Forecasting results on the \textbf{ETTh1} dataset. \(MSE^{\Delta=6}\) reflects the MSE under 6× downsampling. \(576^{\Delta=6}\) and \(336^{\Delta=6}\) correspond to context and prediction sequences with 6× downsampling.}
\label{tab:tech-motivation}
\vspace{-5pt}
\begin{adjustbox}{max width=\linewidth}
\begin{tabular}{lccccc}
\toprule
\textbf{Model} & \textbf{T} & \textbf{H} & \textbf{MSE} & \textbf{MSE$^{\Delta=6}$} \\
\midrule
\multirow{4}{*}{PatchTST}
& 576                 & 336 & 0.480 & 0.477 \\
& \(576^{\Delta=6}\)  & 336 & 0.504 & 0.445 \\
& \cellcolor{blue!20}576  & \cellcolor{blue!20}\(336^{\Delta=6}\) & \cellcolor{blue!20}- & \cellcolor{blue!20}0.564 \\
& \cellcolor{red!20}\(576^{\Delta=6}\)  & \cellcolor{red!20}\(336^{\Delta=6}\) & \cellcolor{red!20}- & \cellcolor{red!20}\textbf{0.433} \\
\midrule
\multirow{4}{*}{DLinear} 
& 576                 & 336 & 0.445 & 0.449 \\
& \(576^{\Delta=6}\)  & 336 & 0.465 & 0.438 \\
& \cellcolor{blue!20}576  & \cellcolor{blue!20}\(336^{\Delta=6}\) & \cellcolor{blue!20}- & \cellcolor{blue!20}0.457 \\
& \cellcolor{red!20}\(576^{\Delta=6}\)  & \cellcolor{red!20}\(336^{\Delta=6}\) & \cellcolor{red!20}- & \cellcolor{red!20}\textbf{0.425} \\
\midrule
\multirow{2}{*}{TimeMixer}
& 576 & 336 & \textbf{0.432} & \textbf{0.435} \\
& 576 & \(336^{\Delta=6}\) & - & 0.461 \\
\bottomrule
\end{tabular}
\end{adjustbox}
\vspace{-10pt}
\end{wraptable}

To understand how scale alignment and temporal granularity affect forecasting performance, we conduct experiments on the ETTh1 dataset with a fixed prediction horizon of \(H = 336\).
As shown in Table~\ref{tab:tech-motivation}, we evaluate models under three input settings: (1) full-resolution input (\(T=576\)); (2) downsampled~\footnote{This operation can be implemented in PyTorch-style code as \texttt{x[:, ::Delta, :]}, where \texttt{x} is the input sequence with dimensions \texttt{[batch\_size, seq\_len, num\_of\_channels]}.}
 input with full-resolution output (\(T=576^{\Delta=6}, H=336\)); and (3) fully scale-aligned input-output (\(T=576^{\Delta=6}, H=336^{\Delta=6}\)). In addition to standard MSE, we report \(\text{MSE}^{\Delta=6}\), which measures prediction quality under 6× temporal downsampling.

We highlight three key observations:

\textbf{\ding{182} Scale misalignment degrades performance.}  
When input and output scales are not aligned (e.g., \(T=576^{\Delta=6}, H=336\)), performance consistently drops compared to the fully aligned setting. This degradation occurs despite the downsampled input offering more context in real time steps, underscoring the importance of maintaining resolution consistency between input and output.

\textbf{\ding{183} Sparse, scale-aligned forecasting improves coarse-scale accuracy.}  
Models trained and evaluated under the same coarse resolution (\(T=576^{\Delta=6}, H=336^{\Delta=6}\)) achieve the lowest \(\text{MSE}^{\Delta=6}\), demonstrating that explicitly modeling forecasting at sparse temporal scales yields more accurate coarse-level predictions—even with significantly fewer input tokens.

\textbf{\ding{184} Multiscale-mixing improves overall accuracy but lacks scale-specific precision.}  
TimeMixer~\citep{wang2024timemixer}, which adopts a multiscale-mixing strategy through grouped temporal operations, achieves the lowest overall MSE. However, its coarse-scale performance (as indicated by \(\text{MSE}^{\Delta=6}\)) remains inferior to that of models trained explicitly at coarse resolutions. This suggests that while multiscale-mixing enhances global modeling capacity, it may obscure scale-specific patterns and reduce fidelity at individual temporal granularities.

These findings motivate our proposed framework, which models forecasting at multiple aligned temporal scales, aiming to preserve both temporal coherence and resolution-specific expressiveness.

\subsection{Modular Scale-wise Autoregressive Framework}
\label{sec:method2}

Real-world time series often exhibit temporal patterns at multiple resolutions—long-term trends unfold gradually, while short-term fluctuations vary rapidly. Modeling such heterogeneous dynamics using a fixed-resolution representation often leads to underfitting or overfitting. To address this, we propose the Modular Scale-wise Autoregressive Framework (MSAR), a model-agnostic pipeline that decomposes forecasting into multiple scale-aligned subproblems. Each scale is handled independently by a dedicated forecasting module, while a lightweight information flow mechanism propagates relevant context from coarse to fine resolutions.

\paragraph{Scale-wise Decoupling and Progressive Forecasting.}
We define a sequence of temporal scales with downsampling factors $\{d_1, d_2, \dots, d_N\}$, where $d_1 > d_2 > \dots > d_N = 1$. Given an input sequence $\mathbf{X}_{T-W:T} \in \mathbb{R}^{W \times C}$ and a forecasting horizon $H$, the scale-$i$ input and target are
\begin{align}
    \mathbf{X}^i = \mathbf{X}^{\Delta=d_i}_{T - W_i:T} \in \mathbb{R}^{\tfrac{W_i}{d_i} \times C}, 
    \mathbf{Y}^i = \mathbf{Y}^{\Delta=d_i}_{1:H}
    \in \mathbb{R}^{\tfrac{H}{d_i} \times C},
\end{align}
where $\mathbf{X}^{\Delta=d}$ denotes a deterministic stride-based sampling operator that retains every $d$-th timestamp without interpolation or filtering, thereby preserving the original temporal structure at a coarser resolution. The $W_i$ is the look-back window at scale $i$.

\paragraph{Forecasting Objective.}  
MSAR formulates multiscale forecasting as a structured autoregressive process:
\begin{equation}
p(\mathbf{Y}^1, \dots, \mathbf{Y}^N \mid \mathbf{X}^1, \dots, \mathbf{X}^N) 
= \prod_{i=1}^N p(\mathbf{Y}^i \mid \mathbf{X}^i, \hat{\mathbf{Y}}^{<i}_\text{align}),
\label{eq:msar}
\end{equation}
where $\hat{\mathbf{Y}}^{<i}_\text{align}$ aggregates coarse-scale predictions
whose time indices coincide with those of scale $i$:
\begin{equation}
    \hat{\mathbf{Y}}^{<i}_\text{align}
    = \bigcup_{j < i}
      \left\{
          \hat{\mathbf{Y}}^j_t
          \,\middle|\,
          t \in \boldsymbol{\tau}^j \cap \boldsymbol{\tau}^i
      \right\}
    \in \mathbb{R}^{\tfrac{H}{d_i} \times C},
\end{equation}
with $\boldsymbol{\tau}^j$ the index set at scale $j$.
For time points in $\boldsymbol{\tau}^i$ that receive no coarse-scale
coverage, we insert zero vectors to match the target size of $\mathbf{Y}^i$:
\begin{equation}
t \notin \bigcup_{j<i}\boldsymbol{\tau}^j
\quad\Rightarrow\quad
\hat{\mathbf{Y}}^{<i}_\text{align}[t] = \mathbf{0}.
\end{equation}

\paragraph{Information Flow Mechanism.}  
The role of each prediction module $\mathcal{F}_\theta^i$ differs by scale. At the coarsest level ($i = 0$), $\mathcal{F}_\theta^0$ acts as a \emph{forecasting model}, generating an initial coarse-resolution prediction solely from the input $\mathbf{X}^0$. For higher scales ($i > 0$), the modules operate as \emph{imputation models}, where each $\mathcal{F}_\theta^i$ refines missing fine-scale values by conditioning on both the current-scale input $\mathbf{X}^i$ and the aligned coarse predictions passed from previous stages. Formally,
\begin{equation}
    \hat{\mathbf{Y}}^i 
    = \mathcal{F}_\theta^i\!\left(\text{concat}(\mathbf{X}^i, \hat{\mathbf{Y}}^{<i}_\text{align})\right),
    \hat{\mathbf{Y}}^i \in \mathbb{R}^{\tfrac{H}{d_i} \times C}.
\end{equation}
This hierarchical design enforces a \emph{scale-aligned information flow}: coarse predictions provide structural guidance, while finer-scale modules progressively enhance temporal resolution through imputation-based refinement.

\paragraph{Refinement in the Imputation Phase.}  
To further enhance cross-scale consistency, MSAR performs complete prediction over the target window $\mathbf{Y}^i$ during the imputation phase. Instead of restricting supervision to only the masked positions ($\text{mask}=0$), we compute the loss across all target positions in the prediction window, including those that were previously filled ($\text{mask}=1$). This approach enables iterative refinement, ensuring that coarse-scale predictions are smoothly integrated and corrected at finer resolutions, while maintaining temporal consistency across scales. This simple yet effective method adjusts previous predictions without introducing any additional parameters, helping to mitigate error accumulation across scales.

\subsection{Training Strategy} 
\label{sec:method3}

We train MSAR using the scale-wise MAE loss:  $\mathcal{L} = \sum_{i=1}^N \text{MAE}(\hat{\mathbf{Y}}^i, \mathbf{Y}^i)$. Directly optimizing this objective with fully autoregressive inputs can lead to slow convergence. During training, finer-scale modules receive ground-truth values from coarser scales, while at inference they receive model predictions. This discrepancy, commonly known as exposure bias, motivates our two-stage training strategy, which gradually transitions from ground-truth conditioning to inference-aligned prediction.

\textbf{Teacher Forcing Training.}  
In the first stage, each scale-specific module $\mathcal{F}_\theta^i$ is trained independently using clean contextual signals. The coarsest forecaster ($i=0$) is optimized directly on ground truth. For finer scales ($i>0$), the model concatenates the ground-truth values of $\mathbf{Y}^{<i}_\text{align}$ from coarser levels with the current input. This teacher-forced approach stabilizes early learning and accelerates convergence by preventing error propagation across scales.  
Importantly, during this phase, the imputation models at finer scales use the ground-truth $Y$ values (rather than previous predictions), ensuring the model learns directly from the real target values without introducing refinement.

\textbf{Joint Training.}  
In the second stage, we switch to inference-aligned inputs: each imputation module receives predictions from coarser levels rather than ground truth. Crucially, during this phase we compute the loss across the forecast horizon at each scale (i.e., all $\tfrac{H}{d_i}$ steps), rather than restricting supervision to unobserved positions. This full-horizon supervision enables progressive refinement: predictions inherited from coarse scales can be corrected at finer resolutions, ensuring temporal coherence across scales. Thanks to its modular and scale-independent design, this curriculum allows MSAR to transition smoothly from robust pretraining under teacher forcing to deployment-consistent autoregressive inference.

\section{Experiments}

\begin{table*}[t]
  \caption{Long-term forecasting with input length 336, averaged over horizons $\{96,192,336,720\}$. Lower is better; bold marks pairwise minima at the reported precision, including ties. Full results: Appendix~\ref{appendix:fullcompare}.}
  \label{tab:long_term_forecasting_results}
  \centering
  \small
  \definecolor{msarTableHeader}{RGB}{231,238,247}
  \setlength{\tabcolsep}{2.8pt}
  \renewcommand{\arraystretch}{0.96}
  \begin{tabular*}{\textwidth}{@{\extracolsep{\fill}\hspace{\tabcolsep}}ll*{5}{cc}@{\hspace{\tabcolsep}}}
    \toprule
    \rowcolor{msarTableHeader}
    & &
    \multicolumn{2}{c}{\textbf{SimpleTM}} & \multicolumn{2}{c}{\textbf{DUET}} &
    \multicolumn{2}{c}{\textbf{iTransformer}} & \multicolumn{2}{c}{\textbf{PatchTST}} &
    \multicolumn{2}{c}{\textbf{DLinear}} \\
    \rowcolor{msarTableHeader}
    \multirow{-2}{*}{\textbf{Dataset}} & \multirow{-2}{*}{\textbf{Metric}} & Base & +MSAR & Base & +MSAR & Base & +MSAR & Base & +MSAR & Base & +MSAR \\
    \midrule
    \multirow{2}{*}{ECL} & MSE & \textbf{0.164} & 0.168 & 0.178 & \textbf{0.177} & 0.167 & \textbf{0.160} & 0.168 & \textbf{0.166} & 0.170 & \textbf{0.168} \\
    & MAE & \textbf{0.254} & 0.259 & 0.277 & \textbf{0.276} & 0.257 & \textbf{0.252} & 0.258 & \textbf{0.254} & 0.269 & \textbf{0.259} \\
    \addlinespace[0.8pt]
    \multirow{2}{*}{ETTh1} & MSE & 0.424 & \textbf{0.410} & \textbf{0.403} & 0.404 & 0.475 & \textbf{0.449} & 0.428 & \textbf{0.413} & 0.420 & \textbf{0.419} \\
    & MAE & 0.437 & \textbf{0.425} & \textbf{0.415} & 0.416 & 0.459 & \textbf{0.455} & 0.439 & \textbf{0.428} & 0.430 & \textbf{0.426} \\
    \addlinespace[0.8pt]
    \multirow{2}{*}{ETTh2} & MSE & 0.377 & \textbf{0.351} & 0.355 & \textbf{0.345} & 0.405 & \textbf{0.395} & 0.404 & \textbf{0.348} & 0.397 & \textbf{0.391} \\
    & MAE & 0.401 & \textbf{0.388} & 0.390 & \textbf{0.384} & 0.420 & \textbf{0.419} & 0.422 & \textbf{0.384} & 0.414 & \textbf{0.412} \\
    \addlinespace[0.8pt]
    \multirow{2}{*}{ETTm1} & MSE & 0.354 & \textbf{0.345} & 0.352 & \textbf{0.351} & 0.388 & \textbf{0.359} & 0.371 & \textbf{0.345} & 0.354 & \textbf{0.353} \\
    & MAE & 0.375 & \textbf{0.371} & 0.370 & \textbf{0.369} & 0.394 & \textbf{0.384} & 0.379 & \textbf{0.369} & 0.370 & \textbf{0.369} \\
    \addlinespace[0.8pt]
    \multirow{2}{*}{ETTm2} & MSE & 0.283 & \textbf{0.256} & \textbf{0.255} & \textbf{0.255} & 0.281 & \textbf{0.273} & 0.258 & \textbf{0.256} & 0.260 & \textbf{0.259} \\
    & MAE & 0.322 & \textbf{0.308} & 0.309 & \textbf{0.308} & 0.335 & \textbf{0.328} & 0.314 & \textbf{0.309} & 0.314 & \textbf{0.313} \\
    \addlinespace[0.8pt]
    \multirow{2}{*}{Exchange} & MSE & 0.438 & \textbf{0.352} & 0.394 & \textbf{0.372} & \textbf{0.403} & 0.437 & 0.465 & \textbf{0.364} & 0.524 & \textbf{0.372} \\
    & MAE & 0.433 & \textbf{0.401} & 0.423 & \textbf{0.408} & \textbf{0.444} & 0.462 & 0.462 & \textbf{0.408} & 0.455 & \textbf{0.407} \\
    \addlinespace[0.8pt]
    \multirow{2}{*}{Traffic} & MSE & \textbf{0.432} & 0.433 & \textbf{0.444} & 0.445 & 0.427 & \textbf{0.419} & 0.426 & \textbf{0.418} & \textbf{0.447} & 0.452 \\
    & MAE & 0.295 & \textbf{0.294} & 0.303 & \textbf{0.302} & 0.292 & \textbf{0.282} & 0.276 & \textbf{0.267} & 0.302 & \textbf{0.278} \\
    \addlinespace[0.8pt]
    \multirow{2}{*}{Weather} & MSE & 0.227 & \textbf{0.226} & 0.248 & \textbf{0.246} & 0.244 & \textbf{0.227} & 0.233 & \textbf{0.230} & 0.246 & \textbf{0.245} \\
    & MAE & 0.254 & \textbf{0.253} & 0.272 & \textbf{0.271} & 0.270 & \textbf{0.262} & 0.256 & \textbf{0.253} & 0.278 & \textbf{0.275} \\
    \addlinespace[0.8pt]
    \multirow{2}{*}{Solar-Energy} & MSE & 0.270 & \textbf{0.242} & 0.260 & \textbf{0.250} & 0.216 & \textbf{0.196} & 0.213 & \textbf{0.206} & \textbf{0.255} & 0.269 \\
    & MAE & 0.292 & \textbf{0.267} & 0.245 & \textbf{0.241} & 0.227 & \textbf{0.225} & 0.232 & \textbf{0.229} & 0.315 & \textbf{0.236} \\
    \addlinespace[0.8pt]
    \multirow{2}{*}{Wind} & MSE & 0.783 & \textbf{0.769} & 0.768 & \textbf{0.767} & 0.780 & \textbf{0.732} & 0.776 & \textbf{0.761} & \textbf{0.749} & 0.750 \\
    & MAE & 0.687 & \textbf{0.677} & \textbf{0.675} & \textbf{0.675} & 0.689 & \textbf{0.664} & 0.682 & \textbf{0.675} & \textbf{0.677} & \textbf{0.677} \\
    \bottomrule
  \end{tabular*}
\end{table*}

We conduct extensive experiments to evaluate the effectiveness of the proposed Modular Scale-wise Autoregressive Framework (MSAR). Our empirical study is designed to address the following key questions: (1). Model-agnostic gains: As a plugin method, can MSAR consistently improve mainstream state-of-the-art (SOTA) forecasting models, and how does it compare against strong baselines? (cf. Sec.\,~\ref{sec:Forecasting_Results}) (2). Efficiency study: What is the training and inference overhead introduced by MSAR compared to vanilla models? (cf. Sec.\,~\ref{sec:Forecasting_Results}) (3). Mechanism analysis: Why does MSAR work? We perform ablation studies to dissect the contribution of its coarse-to-fine autoregressive design. (cf. Sec.\,~\ref{sec:ablation}) (4). Extended context: Can MSAR effectively benefit from longer look-back windows and better exploit historical dependencies? (cf. Sec.\,~\ref{sec:sensitivity}) (5). Sensitivity analysis: How do the number of autoregressive layers and the choice of interval sizes affect forecasting accuracy? (cf. Sec.\,~\ref{sec:sensitivity})

\paragraph{Dataset.}
For the long-term forecasting experiments, we utilize a diverse set of datasets to evaluate the robustness and generalizability of our models across different domains. These datasets include ECL, ETT (4
subsets), Exchange, Traffic, Weather, Solar-Energy, and Wind~\footnote{https://www.kaggle.com/datasets/mubashirrahim/wind-power-generation-data-forecasting}. Additionally, for the short-term forecasting experiments, the PEMS (4 subsets)~\citep{chen2001freeway} dataset is employed, specifically designed for short-term traffic prediction tasks. Following TSLib~\citep{wang2024tssurvey}, we adopt the same dataset splits, data preprocessing steps and ensure no last batch is dropped during training. For long-term forecasting, we use prediction lengths of \{96, 192, 336, 720\}, while for short-term forecasting, we use a prediction length of 12.

\paragraph{Baselines.}
To evaluate the performance of our proposed MSAR framework, we compare it against several state-of-the-art (SOTA) methods, which include SimpleTM~\citep{chen2025simpletm}, DUET~\citep{qiu2025duet}, iTransformer~\citep{liu2024itransformer}, PatchTST~\citep{nie2022time}, and DLinear~\citep{zeng2023transformers}.

\paragraph{Unified experiment settings.}
To fairly evaluate the effectiveness of MSAR, we designed two experimental setups: (1). \textbf{Baseline vs. Baseline + MSAR}: In this experiment, we compare the performance of the baseline models with and without MSAR, ensuring that \textit{all hyperparameters remain consistent across both setups}. (2). \textbf{Full Hyperparameter Search}: For this experiment, we perform a comprehensive search over the full hyperparameter space to ensure that all models are evaluated with the same range of parameter values, allowing for a fair comparison. Detailed experimental settings can be found in the Appendix~\ref{appendix:exp_details}.

\subsection{Forecasting Results}
\label{sec:Forecasting_Results}

\paragraph{Performance Study.}
Table~\ref{tab:long_term_forecasting_results} reports average results across four horizons ${96,192,336,720}$ with all methods extracting features from a look-back window of length 336. Across ten datasets and five backbones, MSAR improves most model--dataset pairs. For instance, DLinear’s MSE on Exchange drops from 0.524 to 0.372, and PatchTST’s MAE on Weather decreases from 0.256 to 0.253. These consistent gains across both lightweight linear models and Transformer-style architectures (PatchTST, iTransformer, DUET, SimpleTM) highlight the generality of MSAR as a plug-in framework. Full per-horizon results are given in Appendix~\ref{appendix:fullcompare}.

We additionally compare methods under a shared hyperparameter search space; the results and analysis are provided in Appendix~\ref{appendix:hparam_results}.

\paragraph{Stratified Performance Analysis.}
To further contextualize the magnitude and consistency of improvements, we conduct a stratified analysis using dataset meta-features reported in TFB~\citep{qiu2024tfb} benchmark, including seasonality strength, trend/non-stationarity, noise ratio, input dimensionality, and effective horizon difficulty. Three observations emerge.

\begin{itemize}[leftmargin=*,itemsep=-0.1em]
\item[\ding{182}] \textbf{Larger gains on datasets with strong periodicity and stable seasonal structure.}  
ETTh1/ETTh2 and ETTm1/ETTm2 are categorized as highly periodic and low-to-moderate noise. MSAR achieves the largest improvements here across nearly all backbones. Coarse scales reliably capture long-range seasonal cycles, providing stable guidance for finer-scale refinement.
\item[\ding{183}] \textbf{MSAR is particularly effective for long horizons where long-range structure dominates.} The strongest improvements appear at horizons 336 and 720 on the ETT series and Wind datasets. This is consistent with the coarse-to-fine design: coarse-scale forecasting models slow-varying patterns, while finer scales reconstruct higher-frequency details.
\item[\ding{184}] \textbf{Moderate but consistent gains on high-noise or weakly periodic datasets.}  
Traffic—characterized by high stochasticity and weak periodicity—shows smaller but still stable improvements. In these settings, coarse-scale signals carry limited structure, and irregular fine-scale fluctuations dominate; MSAR nevertheless improves long-horizon accuracy by stabilizing coarse-level guidance.

\end{itemize}

\begin{figure*}[!t]
\centering
\begin{minipage}[c]{0.46\linewidth}
  \centering
  \captionof{table}{Ablation study on MSAR components over four datasets.
  \ding{172} = Multiscale, \ding{173} = Alignment, \ding{174} = Scale-wise AR.
  Results with $^{\ast}$: no joint training; $^{\dagger}$: no full pred-len imputation loss.}
  \label{tab:ablation_msar_full}
  \renewcommand{\arraystretch}{1.0}
  \setlength{\tabcolsep}{2pt}
  \tiny
  \begin{threeparttable}
  \begin{tabular}{c|cc|cc|cc|cc}
    \Xhline{1pt}
    \rowcolor[rgb]{0.867,0.875,0.91}
      & \multicolumn{2}{c|}{ECL}
      & \multicolumn{2}{c|}{ETTm1}
      & \multicolumn{2}{c|}{Traffic}
      & \multicolumn{2}{c}{Weather} \\
    \cline{2-9}
    \multirow{-2}{*}{\cellcolor[rgb]{0.867,0.875,0.91}\scalebox{0.85}{Setting}}
      & \cellcolor[rgb]{0.867,0.875,0.91}\scalebox{0.75}{MSE}
      & \cellcolor[rgb]{0.867,0.875,0.91}\scalebox{0.75}{MAE}
      & \cellcolor[rgb]{0.867,0.875,0.91}\scalebox{0.75}{MSE}
      & \cellcolor[rgb]{0.867,0.875,0.91}\scalebox{0.75}{MAE}
      & \cellcolor[rgb]{0.867,0.875,0.91}\scalebox{0.75}{MSE}
      & \cellcolor[rgb]{0.867,0.875,0.91}\scalebox{0.75}{MAE}
      & \cellcolor[rgb]{0.867,0.875,0.91}\scalebox{0.75}{MSE}
      & \cellcolor[rgb]{0.867,0.875,0.91}\scalebox{0.75}{MAE} \\
    \Xhline{1pt}

    \ding{172}
      & 0.199 & 0.290
      & 0.383 & 0.395
      & 0.472 & 0.306
      & 0.255 & 0.277 \\

    \ding{173}
      & 0.205 & 0.292
      & 0.387 & 0.398
      & 0.486 & 0.314
      & 0.264 & 0.283 \\

    \ding{174}
      & 0.212 & 0.298
      & 0.395 & 0.405
      & 0.495 & 0.322
      & 0.271 & 0.289 \\

    \ding{172}+\ding{173}
      & 0.184 & 0.277
      & 0.379 & 0.389
      & 0.456 & 0.293
      & 0.244 & 0.269 \\

    \ding{172}+\ding{174}
      & 0.191 & 0.283
      & 0.381 & 0.393
      & 0.460 & 0.297
      & 0.247 & 0.272 \\

    \rowcolor{gray!20}
    \ding{172}+\ding{173}+\ding{174}
      & \textbf{0.172} & \textbf{0.263}
      & \textbf{0.355} & \textbf{0.374}
      & \textbf{0.441} & \textbf{0.280}
      & \textbf{0.234} & \textbf{0.253} \\

    \rowcolor{gray!20}
    \ding{172}+\ding{173}+\ding{174}$^{\ast}$
      & 0.181 & 0.273
      & 0.375 & 0.386
      & 0.451 & 0.290
      & 0.241 & 0.267 \\

    \rowcolor{gray!20}
    \ding{172}+\ding{173}+\ding{174}$^{\dagger}$
      & 0.182 & 0.276
      & 0.366 & 0.383
      & 0.444 & 0.287
      & 0.237 & 0.264 \\

    \bottomrule
  \end{tabular}
  \end{threeparttable}
\end{minipage}
\hfill
\begin{minipage}[c]{0.52\linewidth}
    \centering
    \includegraphics[width=\linewidth]{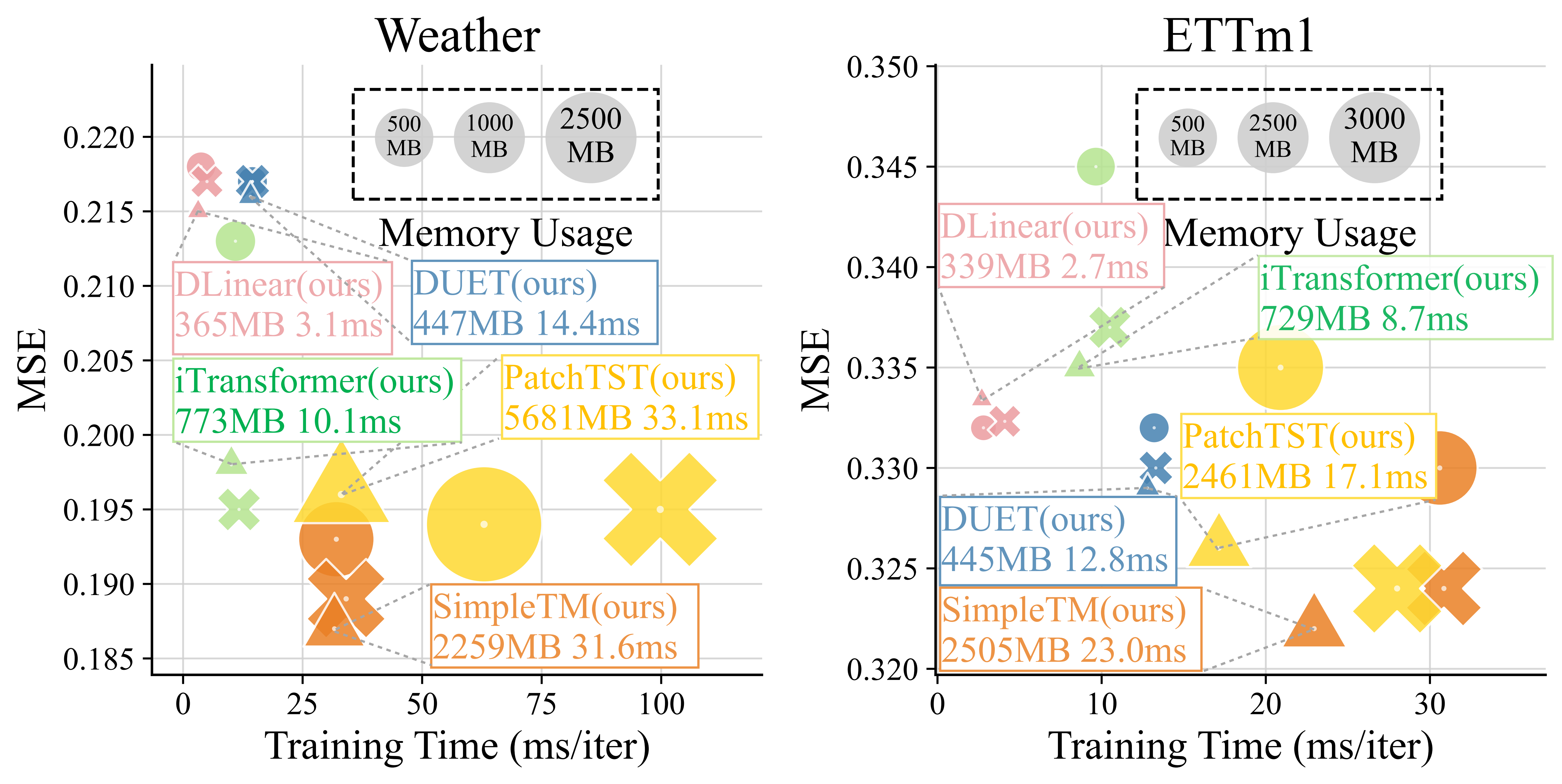}
    \captionof{figure}{Efficiency–accuracy trade-off on \textbf{Weather} (21 variates) and \textbf{ETTm1} (7 variates). Different symbols indicate model variants: circles (\gfcircle) represent the baseline, crosses (\gfcross) represent the full version of MSAR, and triangles (\gftriangle) represent the flexible version of MSAR.}
    \label{fig:parameters}
\end{minipage}
\end{figure*}

\paragraph{Training Efficiency.}
Beyond forecasting accuracy, we further evaluate the \emph{efficiency} of MSAR. 
Figure~\ref{fig:parameters} plots training time (x-axis), error (y-axis), and GPU memory (bubble size) for five representative baselines and their MSAR-augmented counterparts on Weather and ETTm1 ($H{=}336$, $T{=}192$). 
We report two variants of MSAR: \textit{full} and \textit{flexible}. 

The \textit{full} variant applies a unified look-back length of $336$ for both forecasting and imputation. 
In this case, the forecasting module consumes a compressed $56$-length input (via $6\times$ downsampling), while the imputation module operates on the full $336$ tokens.  
The \textit{flexible} variant instead shortens the imputation look-back to $192$, exploiting MSAR’s plug-in, model-agnostic design. 
This simple change leads to a substantial reduction in training cost while often \emph{improving} accuracy.

MSAR incurs additional training-time overhead.
Although each iteration is lighter (lower ms/iter) due to scale decoupling, the use of a two-stage training strategy to mitigate teacher-forcing-induced train--test discrepancy leads to about \(1.67\times\) longer wall-clock time than single-stage baselines.
This cost is offset by more stable convergence, improved inference behavior, and reduced memory usage from decoupled per-scale forecasters, yielding a favorable efficiency--accuracy trade-off across diverse backbones.

\paragraph{Inference Efficiency.}
We further evaluate inference-time efficiency in Appendix~\ref{appendix:efficiency}. MSAR reduces inference latency for PatchTST~\citep{nie2022time}, DLinear~\citep{zeng2023transformers}, and TimesNet~\citep{wu2022timesnet}—often by nearly a factor of two—because scale-wise forecasting shortens the effective sequence length processed at the fine scale. For iTransformer~\citep{liu2024itransformer}, MSAR introduces only a slight overhead, which is expected given its variate-wise embedding over the full sequence, where reducing token count does not yield proportional savings. 

\subsection{Ablation Study}
\label{sec:ablation}

In this section, we ask: \textit{where does the performance gain of MSAR come from?} We ablate three design components: \ding{172} Multiscale, \ding{173} Alignment modeling in encoder, and \ding{174} Scale-wise Autoregression (AR), using PatchTST~\citep{nie2022time} as the backbone. Alignment enforces cross-scale consistency between the input \(X\) and output \(Y\).

Table~\ref{tab:ablation_msar_full} shows that no single component alone is sufficient.
Using only \ding{172} outperforms \ding{173} and \ding{174}, indicating the benefit of multi-resolution context, but its gains are limited by scale mixing.
\ding{173} behaves similarly to a single-scale baseline, while \ding{174} yields the weakest performance due to the lack of proper input--output alignment.
Pairwise combinations improve performance, with \ding{172}+\ding{173} consistently outperforming \ding{172}+\ding{174}, highlighting the importance of alignment in mitigating multiscale interference.

When all components are combined (\ding{172}+\ding{173}+\ding{174}), MSAR achieves the best results across all datasets.
The full design removes encoder-side scale mixing, preserves strict input--output alignment, and enables clean scale-wise autoregressive refinement.
The consistent gains over all ablated variants confirm that MSAR’s improvements stem from the \textbf{synergistic interaction} of all three components rather than any single one.

\paragraph{Joint training and full-horizon supervision.}
The last two rows of Table~\ref{tab:ablation_msar_full} isolate the training choices from the architectural components. On ETTm1, removing joint training increases MSE from $0.355$ to $0.375$, while removing the full prediction-window imputation loss increases it to $0.366$. Both changes also degrade MSE and MAE on ECL, Traffic, and Weather. These comparisons support two complementary roles: joint training exposes the refinement modules to predicted coarse-scale context, while full-horizon supervision allows fine-scale modules to correct inherited predictions as well as fill missing positions. The gains therefore depend on inference-aligned conditioning and supervision, not merely on adding scales or layers.

\subsection{Model Analysis}

\begin{figure*}[!t]
    \centering
    \includegraphics[width=0.95\linewidth,height=0.4874803103\linewidth]{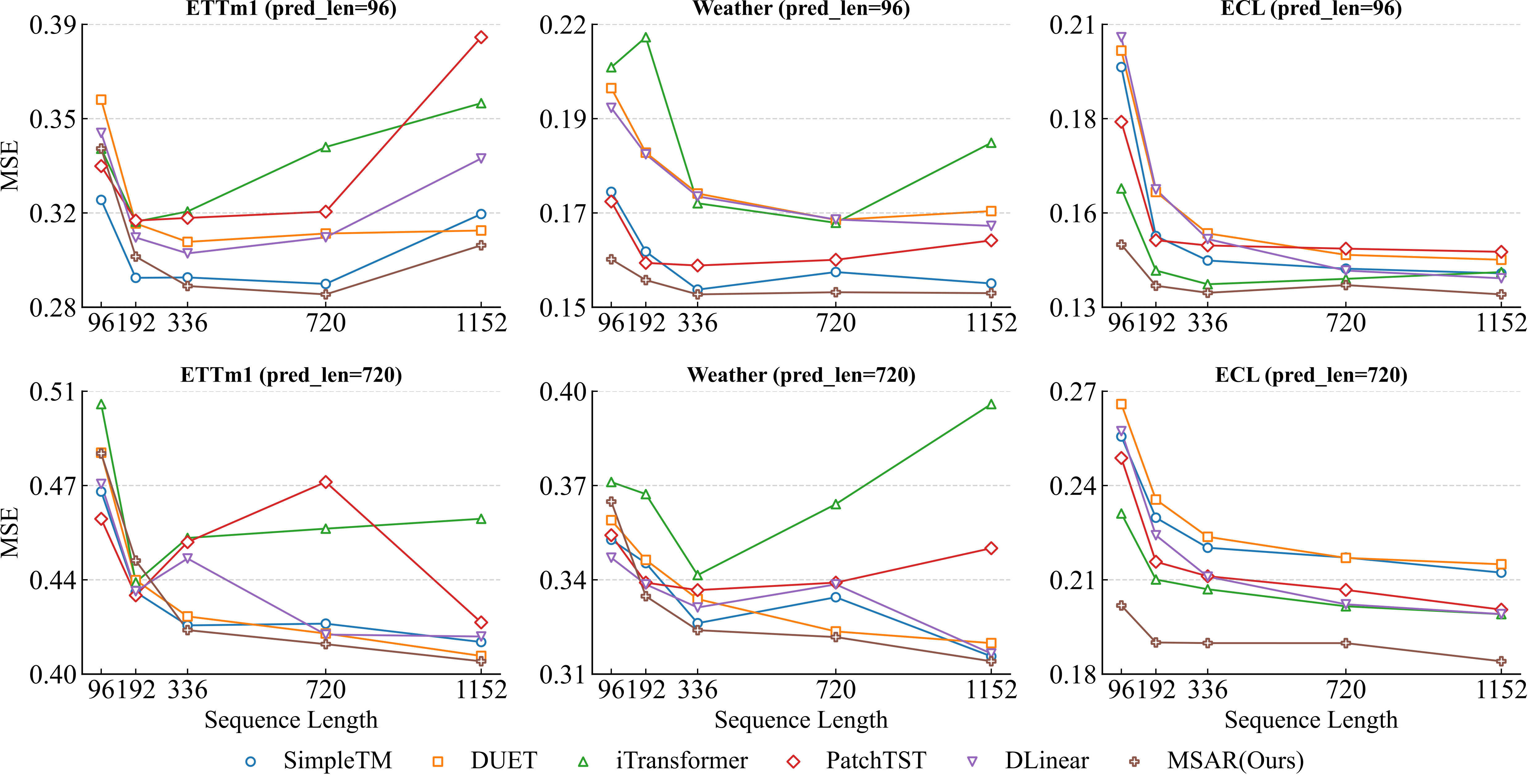}
    \caption{Impact of look-back window length on forecasting performance (measured by MSE).
    We report results on three representative datasets (ETTm1, Weather, and Electricity) under two horizons ($H=96$ and $H=720$).
    Each curve shows the error as the input sequence length increases from $96$ to $1152$.}
    \label{fig:lookbackwindow}
\end{figure*}

\paragraph{Extended context.}
We vary the look-back window $T\in\{96,192,336,720,1152\}$ on ETTm1, Weather, and Electricity under two prediction horizons ($H=96$ and $H=720$). Figure~\ref{fig:lookbackwindow} compares MSAR with five baselines. Longer histories generally reduce MSAR's error, with substantial gains when extending the context from 96 to 336 steps. At $T=1152$, MSAR remains among the strongest methods in all six dataset--horizon settings, showing that the framework can benefit from historical information beyond the fixed input length used in Table~\ref{tab:long_term_forecasting_results}.

The benefit is not uniformly monotonic. On ETTm1 with $H=96$, MSAR's error increases from $T=720$ to $1152$, whereas its long-horizon error continues to decrease. Electricity shows diminishing gains after the initial context extension, and several baselines fluctuate on ETTm1 and Weather. Thus, a longer history is useful only when the model can extract relevant structure; context length should still be selected for each dataset and horizon.

\Needspace{24\baselineskip}
\begin{wrapfigure}[20]{r}{0.50\textwidth}
    \vspace{-4pt}
    \centering
    \includegraphics[width=\linewidth,height=0.8909090909\linewidth]{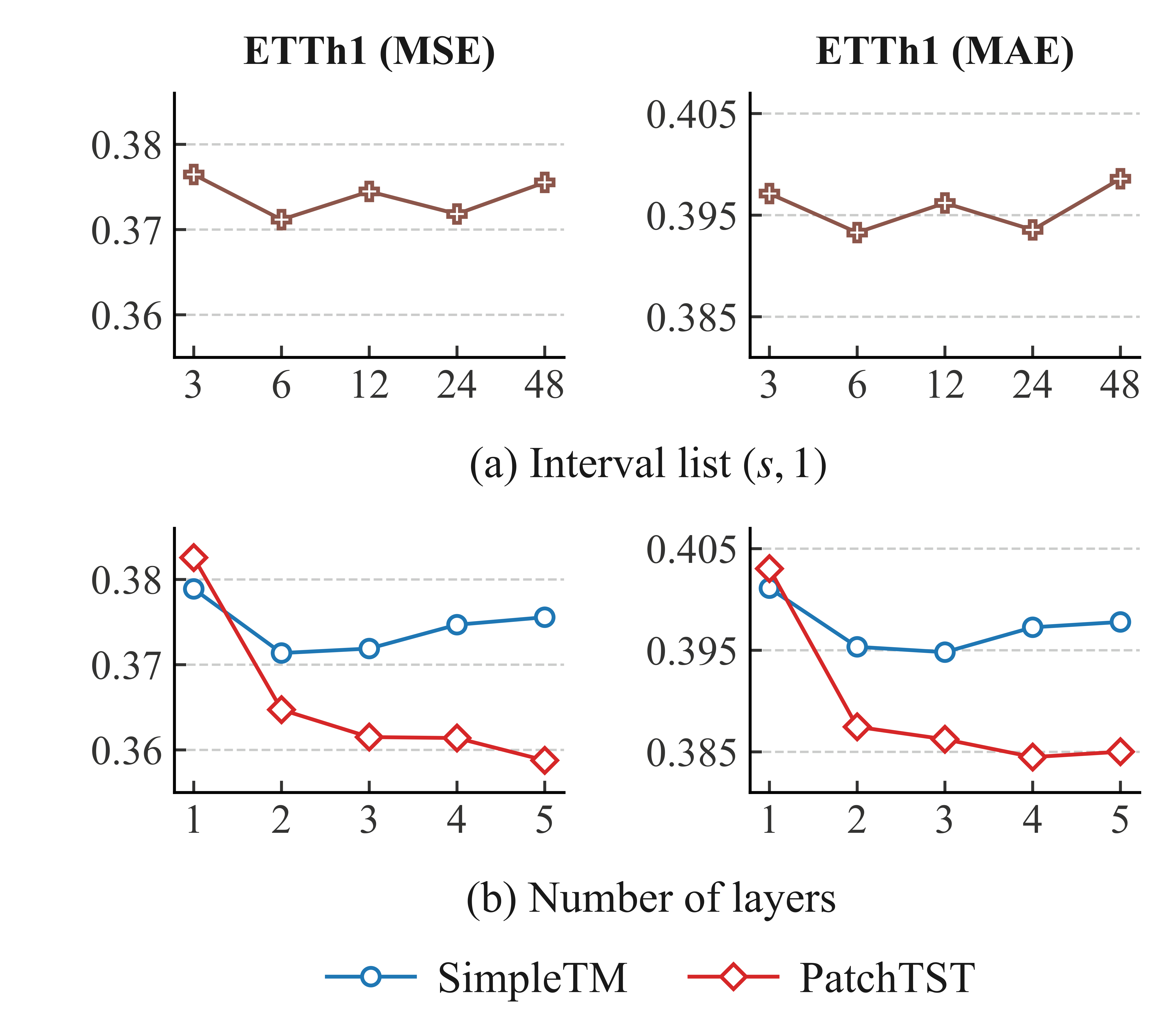}
    \captionsetup{font=small,skip=3pt}
    \caption{MSAR sensitivity on ETTh1; lower is better.
    \textbf{(a)} Interval lists $(s,1)$, averaged over five backbones.
    \textbf{(b)} Layer counts for PatchTST and SimpleTM; one layer is the baseline.
    Axis limits are shared within each metric.}
    \label{fig:sensitive}
    \vspace{-6pt}
\end{wrapfigure}

\paragraph{Sensitivity Analysis.}
\label{sec:sensitivity}
We further investigate the sensitivity of MSAR with respect to the choice of interval list and the number of layers on the ETTh1 dataset. As shown in Figure~\ref{fig:sensitive}(a), the top row reports the average performance of five backbones under different interval lists, with MSE and MAE shown separately. The results remain stable across varying downsampling factors, indicating that MSAR is largely insensitive to the specific interval choices and thus requires minimal tuning. Figure~\ref{fig:sensitive}(b) illustrates the impact of the number of layers. While adding one or two imputation layers consistently improves both MSE and MAE over the baseline, further increasing the number of layers yields diminishing returns and in some cases even slight degradation. These observations confirm that MSAR achieves robust performance across settings, with a favorable trade-off between accuracy and complexity.

\paragraph{Full-horizon refinement.}
Full-horizon supervision allows fine-scale modules to revise inherited predictions rather than treating them as fixed outputs. Coarse predictions serve as conditioning inputs, while each finer module predicts the entire target window using its higher-resolution history. Joint training exposes this correction process to predicted context instead of exact target values. Together, these choices provide a mechanism for limiting inherited errors, supported by the ablations above, without guaranteeing that errors disappear as more scales are added.

\paragraph{Interaction with channel modeling.}
MSAR applies the same temporal stride to every channel, retaining all $C$ variables at each selected time point. Scale construction neither partitions nor mixes variables, so each backbone retains its own channel-modeling mechanism. Channel-independent backbones such as PatchTST and DLinear remain channel independent, while iTransformer and DUET can model cross-channel relationships within each scale. This distinction concerns the architecture: subsampling changes the temporal information available at coarse scales and need not preserve every cross-channel dependency. Finer modules also receive their own higher-resolution historical inputs, allowing them to use information absent from coarse predictions. MSAR therefore complements the backbone's variable modeling through temporal scale alignment.

\paragraph{Limitations and Future Work.}
MSAR can propagate coarse-scale errors, adds training stages, and does not improve every backbone--dataset setting. The reported improvements are based on point estimates, and we do not systematically characterize variability across random seeds. The flexible variant reduces computation, but real-time deployment may require adaptive scheduling or scale selection. Future work includes joint multiscale pretraining and scaling studies, residual corrections to inherited predictions, and refinement under structured noise~\citep{chen2024diffusion} to improve robustness to imperfect coarse context.

\section{Conclusions}

In this study, we propose the Modular Scale-wise Autoregressive Framework (MSAR) to overcome the limitations of existing time series forecasting methods that suffer from scale misalignment and limited long-context utilization. Unlike prior approaches that apply multiscale modeling only to the input sequence or rely on single-scale decoding, MSAR performs fully scale-aligned modeling across both inputs and outputs, enabling each scale to be modeled independently. By introducing a scale-wise autoregressive strategy, MSAR progressively refines predictions from coarse to fine resolutions through an efficient, modular information flow. Our work highlights the importance of aligning temporal granularity in both modeling and generation, paving the way for scalable and robust multiscale forecasting.

\clearpage
\begin{ack}
\paragraph{Funding.}
This work was supported by the Innovation Funding of ICT, CAS under Grant No.~E661110000. Additional support was provided by the Key Scientific Research Project of Higher Education Institutions of Henan Province (No.~24A520006) and the Research Fund of the School of Computer and Information Engineering (No.~K24061Y), both awarded to Nana Wang.

\paragraph{Competing Interests.}
The authors declare no competing interests.
\end{ack}

\appendix

\section{Implementation Details}

\subsection{Dataset Descriptions}\label{sec:dataset}

We conduct experiments on \textbf{fourteen} real-world multivariate time series datasets, including ten \textbf{long-term} forecasting datasets and four \textbf{short-term} forecasting datasets. The datasets span diverse domains including electricity, weather, traffic, and renewable energy, providing a range of settings for evaluating our approach. Table~\ref{tab:datasets} summarizes the statistics of all datasets. Specifically:

\begin{itemize}
    \item \textbf{Electricity Transformer Temperature (ETT)}~\cite{haoyietal-informer-2021}: Four subsets are provided. ETTh1/ETTh2 are hourly, while ETTm1/ETTm2 are recorded every 15 minutes. All series are collected from two transformers.
    
    \item \textbf{Weather}~\cite{wu2021autoformer}: 21 meteorological indicators collected every 10 minutes in Germany during 2020 from the Max Planck Biogeochemistry Institute.
    
    \item \textbf{Traffic}~\cite{wu2021autoformer}: Hourly road occupancy rates from 862 sensors on San Francisco Bay Area freeways (2015–2016).
    
    \item \textbf{Electricity}~\cite{wu2021autoformer}: Hourly consumption of 321 clients from 2012–2014.
    
    \item \textbf{Exchange-rate}~\cite{wu2021autoformer}: Daily exchange rates of 8 countries from 1990–2016.
    
    \item \textbf{Solar-energy}~\cite{LSTNet}: 10-minute solar power production from 137 photovoltaic plants in 2006.
    
    \item \textbf{Wind}\footnote{https://www.kaggle.com/datasets/mubashirrahim/wind-power-generation-data-forecasting}: Hourly wind speed and power generation from four real-world wind farms.
    
    \item \textbf{PEMS}: High-frequency traffic measurements with 5-minute resolution. We use four subsets (PEMS03, PEMS04, PEMS07, and PEMS08), each corresponding to different spatial sensor networks.
\end{itemize}

\begin{table}[th]
    \caption{Dataset statistics. ``Dim'' denotes the number of variates; ``Dataset Size'' reports benchmark (Train/Validation/Test) split sizes; ``Frequency'' is the sampling interval. PEMS split sizes follow iTransformer~\citep{liu2024itransformer}.}
  \label{tab:datasets}
  \centering
  \setlength\tabcolsep{3pt}
  \begin{tabular}{c|c|c|c}
    \hline
    Dataset & Dim & Dataset Size (Train/Val/Test) & Frequency \\
    \hline
    ETTh1, ETTh2 & 7 & (8545, 2881, 2881) & Hourly \\
    \hline
    ETTm1, ETTm2 & 7 & (34465, 11521, 11521) & 15 min \\
    \hline
    Weather & 21 & (36792, 5271, 10540) & 10 min \\
    \hline
    Traffic & 862 & (12185, 1757, 3509) & Hourly \\
    \hline
    Electricity & 321 & (18317, 2633, 5261) & Hourly \\
    \hline
    Exchange-rate & 8 & (5120, 665, 1422) & Daily \\
    \hline
    Solar-energy & 137 & (36792, 5271, 10540) & 10 min \\
    \hline
    Wind & 9 & (30468, 4286, 8665) & Hourly \\
    \hline
    PEMS03 & 358 & (15617, 5135, 5135) & 5 min \\
    \hline
    PEMS04 & 307 & (10172, 3375, 3375) & 5 min \\
    \hline
    PEMS07 & 883 & (16911, 5622, 5622) & 5 min \\
    \hline
    PEMS08 & 170 & (10690, 3548, 3548) & 5 min \\
    \hline
  \end{tabular}
\end{table}

\subsection{Evaluation Metrics}
\label{sec:metrics}

We follow standard practices in time series forecasting and adopt different metrics for short-term and long-term datasets.

\textbf{Short-term forecasting.}  
For short-term benchmarks such as PEMS, we evaluate models using three complementary error measures:  
\begin{itemize}
    \item Mean Absolute Error (MAE): $\tfrac{1}{T}\sum_{t=1}^T |\hat{y}_t - y_t|$  
    \item Mean Absolute Percentage Error (MAPE): $\tfrac{1}{T}\sum_{t=1}^T \tfrac{|\hat{y}_t - y_t|}{|y_t|}$  
    \item Root Mean Squared Error (RMSE): $\sqrt{\tfrac{1}{T}\sum_{t=1}^T (\hat{y}_t - y_t)^2}$  
\end{itemize}
MAE reflects absolute deviation, MAPE measures relative deviation with respect to ground truth magnitude, and RMSE penalizes large errors more heavily. Together, these metrics provide a comprehensive evaluation of accuracy and robustness.

\textbf{Long-term forecasting.}  
For long-term benchmarks such as ETT, Weather, and Exchange, we report Mean Squared Error (MSE) and Mean Absolute Error (MAE):  
\begin{itemize}
    \item Mean Squared Error (MSE): $\tfrac{1}{T}\sum_{t=1}^T (\hat{y}_t - y_t)^2$  
    \item Mean Absolute Error (MAE): $\tfrac{1}{T}\sum_{t=1}^T |\hat{y}_t - y_t|$  
\end{itemize}
These metrics are widely adopted in prior work~\cite{nie2022time}, and capture both variance-sensitive and absolute error perspectives. MSE emphasizes stability for long horizons by penalizing large deviations, while MAE provides scale-consistent interpretability. All metrics are averaged across all variates and prediction horizons. Lower values indicate better forecasting performance.

\subsection{Experiment details}
\label{appendix:exp_details}
\label{appendix:A3}
To fairly evaluate the effectiveness of MSAR, we design two experimental setups:
(1) \textbf{Baseline vs. Baseline+MSAR}: we compare the performance of each baseline model with and without MSAR, while keeping \textit{all hyperparameters consistent} across both settings. 
(2) \textbf{Full Hyperparameter Search}: we perform a comprehensive grid search over the full hyperparameter space so that all models are tuned under the same parameter search range.

\paragraph{Settings for (1).}
For the baseline vs. baseline+MSAR setup, we adopt a unified input length (\texttt{seq\_len} = 336) across all datasets, with the detailed hyperparameters summarized in Table~\ref{tab:model_config}.
For MSAR, we fix the interval list to $\texttt{interval\_list} = [6, 1]$, which corresponds to $\texttt{seqlen\_list} = [56, 336]$ for coarse- and fine-scale autoregression. 
This configuration guarantees that the baseline models and their MSAR-augmented counterparts are evaluated under identical input conditions, thereby ensuring a fair comparison.

\paragraph{Setup for (2).}
\label{sec:hparam_search}
For the full hyperparameter search, all methods are tuned under an identical search space to ensure fairness.  
Specifically, the candidate ranges are:
\begin{itemize}
    \item \texttt{seq\_len} $\in \{96, 192, 320, 512\}$, 
    \item \texttt{d\_model} $\in \{16, 128, 512\}$,
    \item \texttt{e\_layers} $\in \{1, 3, 5\}$,
    \item \texttt{epoch} $\in \{50\}$,
    \item learning rate $\in \{1\!\times\!10^{-5}, 1\!\times\!10^{-4}, 1\!\times\!10^{-3}, 1\!\times\!10^{-2}, 0.05\}$.
\end{itemize}
Other hyperparameters are fixed across all models, including \texttt{batch\_size}$=32$, \texttt{n\_heads}$=8$, and \texttt{d\_layers}$=1$.  

This unified search space covers model width, depth, horizon length, and optimization settings, and is applied consistently across all baselines.  
Therefore, the reported results reflect the intrinsic modeling capability of each method rather than differences in hyperparameter tuning.

All experiments are implemented in Pytorch \cite{paszke2017automatic}, and conducted on two machines, each equipped with four NVIDIA 4090 GPUs.

\begin{table}[t]
  \caption{Experiment configuration of MSAR in Table~\ref{tab:base_msar}. All experiments use the Adam optimizer with a learning rate of $10^{-3}$ and early stopping (patience = 3).}
  \label{tab:model_config}
  \vskip 0.05in
  \centering
  \begin{threeparttable}
  \begin{small}
  \renewcommand{\multirowsetup}{\centering}
  \setlength{\tabcolsep}{4.3pt}
  \begin{tabular}{c|c|c|c|c|c|c|c|c}
    \toprule
    \multirow{2}{*}{Dataset / Configurations} & \multicolumn{4}{c}{Model Hyper-parameter} & \multicolumn{4}{c}{Training Process} \\
    \cmidrule(lr){2-5} \cmidrule(lr){6-9}
    & $d_{model}$ & $d_{ff}$ & \ Layers & \ Heads  & Batch Size & Epochs & LR$^\ast$  & Patience\\
    \toprule
    ECL      & 128 & 256  & 1 & 2 & 16 & 10 & $10^{-3}$ & 3 \\
    \midrule
    Traffic  & 128 & 256  & 1 & 2 & 16 & 10 & $10^{-3}$ & 3 \\
    \midrule
    Solar    & 128 & 256  & 1 & 2 & 16 & 10 & $10^{-3}$ & 3 \\
    \midrule
    ETTh1    & 512 & 2048 & 2 & 8 & 32 & 10 & $10^{-3}$ & 3 \\
    \midrule
    ETTh2    & 512 & 2048 & 2 & 8 & 32 & 10 & $10^{-3}$ & 3 \\
    \midrule
    ETTm1    & 512 & 2048 & 2 & 8 & 32 & 10 & $10^{-3}$ & 3 \\
    \midrule
    ETTm2    & 512 & 2048 & 2 & 8 & 32 & 10 & $10^{-3}$ & 3 \\
    \midrule
    Weather  & 512 & 2048 & 2 & 8 & 32 & 10 & $10^{-3}$ & 3 \\
    \midrule
    Exchange & 512 & 2048 & 2 & 8 & 32 & 10 & $10^{-3}$ & 3 \\
    \midrule
    Wind     & 512 & 2048 & 2 & 8 & 32 & 10 & $10^{-3}$ & 3 \\
    \bottomrule
    \end{tabular}
    \begin{tablenotes}
        \footnotesize
        \item[] $\ast$ LR means the initial learning rate.
  \end{tablenotes}
    \end{small}
  \end{threeparttable}
\end{table}

\section{More Results}

\paragraph{Full Results in Table~\ref{tab:long_term_forecasting_results}.}
\label{appendix:fullcompare}
Table~\ref{tab:base_msar} reports the \emph{full} long-term forecasting results across all prediction lengths $\{96, 192, 336, 720\}$ on the benchmark datasets. Here we provide the complete results to ensure full transparency and allow a more fine-grained comparison across different forecasting lengths.  

\begin{table}[t]
\caption{
Full long-term forecasting results on 10 datasets. Avg rows average the four displayed horizons, rounded to three decimals; ties at this precision are both bold.}
\label{tab:base_msar}
\centering
\renewcommand{\arraystretch}{1.15}
\setlength{\tabcolsep}{4pt}
\resizebox{0.95\linewidth}{!}{\begin{tabular}{c c|cccc|cccc|cccc|cccc|cccc}
\toprule
\multirow{3}{*}{Models} & \multirow{3}{*}{Metric} 
& \multicolumn{4}{c}{SimpleTM} 
& \multicolumn{4}{c}{DUET} 
& \multicolumn{4}{c}{iTransformer} 
& \multicolumn{4}{c}{PatchTST} 
& \multicolumn{4}{c}{DLinear} \\
\cmidrule(lr){3-6} \cmidrule(lr){7-10} \cmidrule(lr){11-14} \cmidrule(lr){15-18} \cmidrule(lr){19-22}
& & \multicolumn{2}{c}{Base} & \multicolumn{2}{c}{+Ours} 
& \multicolumn{2}{c}{Base} & \multicolumn{2}{c}{+Ours} 
& \multicolumn{2}{c}{Base} & \multicolumn{2}{c}{+Ours} 
& \multicolumn{2}{c}{Base} & \multicolumn{2}{c}{+Ours} 
& \multicolumn{2}{c}{Base} & \multicolumn{2}{c}{+Ours} \\
\cmidrule(lr){3-4} \cmidrule(lr){5-6} 
\cmidrule(lr){7-8} \cmidrule(lr){9-10} 
\cmidrule(lr){11-12} \cmidrule(lr){13-14} 
\cmidrule(lr){15-16} \cmidrule(lr){17-18} 
\cmidrule(lr){19-20} \cmidrule(lr){21-22}
& & MSE & MAE & MSE & MAE 
& MSE & MAE & MSE & MAE 
& MSE & MAE & MSE & MAE 
& MSE & MAE & MSE & MAE 
& MSE & MAE & MSE & MAE \\
\midrule

\multirow{5}{*}{\rotatebox[origin=c]{90}{Electricity}} 
& 96  & \textbf{0.134} & \textbf{0.226} & 0.137 & 0.231 & 0.149 & 0.251 & \textbf{0.147} & \textbf{0.250} & 0.135 & 0.229 & \textbf{0.132} & \textbf{0.226} & 0.138 & 0.231 & \textbf{0.136} & \textbf{0.227} & 0.143 & 0.242 & \textbf{0.141} & \textbf{0.234} \\
& 192  & 0.154 & 0.245 & \textbf{0.152} & \textbf{0.244} & 0.163 & 0.264 & \textbf{0.162} & \textbf{0.263} & 0.153 & 0.245 & \textbf{0.149} & \textbf{0.241} & 0.154 & 0.246 & \textbf{0.152} & \textbf{0.241} & 0.157 & 0.255 & \textbf{0.155} & \textbf{0.246} \\
& 336  & 0.175 & 0.268 & \textbf{0.167} & \textbf{0.258} & 0.181 & 0.281 & \textbf{0.179} & \textbf{0.279} & 0.169 & 0.261 & \textbf{0.164} & \textbf{0.257} & 0.171 & 0.261 & \textbf{0.168} & \textbf{0.257} & 0.172 & 0.272 & \textbf{0.170} & \textbf{0.262} \\
& 720  & \textbf{0.191} & \textbf{0.278} & 0.214 & 0.303 & 0.220 & 0.311 & \textbf{0.219} & \textbf{0.310} & 0.209 & 0.294 & \textbf{0.196} & \textbf{0.285} & 0.210 & 0.294 & \textbf{0.207} & \textbf{0.290} & 0.207 & 0.305 & \textbf{0.204} & \textbf{0.293} \\
\cline{2-22}
& \textbf{Avg}& \textbf{0.164} & \textbf{0.254} & 0.168 & 0.259 & 0.178 & 0.277 & \textbf{0.177} & \textbf{0.276} & 0.167 & 0.257 & \textbf{0.160} & \textbf{0.252} & 0.168 & 0.258 & \textbf{0.166} & \textbf{0.254} & 0.170 & 0.269 & \textbf{0.168} & \textbf{0.259} \\

\midrule

\multirow{5}{*}{\rotatebox[origin=c]{90}{ETTm1}} 
& 96  & 0.291 & 0.337 & \textbf{0.282} & \textbf{0.330} & \textbf{0.292} & 0.335 & 0.293 & \textbf{0.335} & 0.318 & 0.355 & \textbf{0.300} & \textbf{0.344} & 0.317 & 0.343 & \textbf{0.280} & \textbf{0.327} & 0.293 & 0.333 & \textbf{0.293} & \textbf{0.333} \\
& 192  & 0.330 & 0.363 & \textbf{0.324} & \textbf{0.357} & 0.332 & 0.359 & \textbf{0.330} & \textbf{0.356} & 0.370 & 0.383 & \textbf{0.337} & \textbf{0.370} & 0.335 & 0.361 & \textbf{0.324} & \textbf{0.356} & 0.332 & 0.355 & \textbf{0.332} & \textbf{0.355} \\
& 336  & 0.369 & 0.384 & \textbf{0.357} & \textbf{0.381} & 0.365 & 0.378 & \textbf{0.363} & \textbf{0.376} & 0.400 & 0.403 & \textbf{0.372} & \textbf{0.393} & 0.381 & 0.384 & \textbf{0.361} & \textbf{0.379} & 0.366 & \textbf{0.376} & \textbf{0.366} & 0.377 \\
& 720  & 0.425 & 0.416 & \textbf{0.416} & \textbf{0.415} & 0.419 & 0.409 & \textbf{0.419} & \textbf{0.409} & 0.462 & 0.434 & \textbf{0.428} & \textbf{0.427} & 0.451 & 0.427 & \textbf{0.414} & \textbf{0.412} & 0.426 & 0.417 & \textbf{0.422} & \textbf{0.411} \\
\cline{2-22}
& \textbf{Avg}& 0.354 & 0.375 & \textbf{0.345} & \textbf{0.371} & 0.352 & 0.370 & \textbf{0.351} & \textbf{0.369} & 0.388 & 0.394 & \textbf{0.359} & \textbf{0.384} & 0.371 & 0.379 & \textbf{0.345} & \textbf{0.369} & 0.354 & 0.370 & \textbf{0.353} & \textbf{0.369} \\

\midrule

\multirow{5}{*}{\rotatebox[origin=c]{90}{ETTm2}} 
& 96  & 0.177 & 0.255 & \textbf{0.165} & \textbf{0.246} & 0.163 & 0.248 & \textbf{0.163} & \textbf{0.247} & 0.179 & 0.266 & \textbf{0.172} & \textbf{0.255} & 0.173 & 0.258 & \textbf{0.164} & \textbf{0.246} & 0.165 & 0.248 & \textbf{0.165} & \textbf{0.247} \\
& 192  & 0.251 & 0.302 & \textbf{0.221} & \textbf{0.287} & 0.219 & 0.285 & \textbf{0.219} & \textbf{0.285} & 0.254 & 0.323 & \textbf{0.228} & \textbf{0.298} & \textbf{0.219} & 0.288 & 0.224 & \textbf{0.288} & 0.222 & 0.288 & \textbf{0.222} & \textbf{0.288} \\
& 336  & 0.305 & 0.338 & \textbf{0.272} & \textbf{0.319} & 0.273 & 0.322 & \textbf{0.272} & \textbf{0.322} & 0.300 & \textbf{0.350} & \textbf{0.296} & 0.351 & 0.279 & 0.327 & \textbf{0.279} & \textbf{0.324} & 0.276 & 0.327 & \textbf{0.276} & \textbf{0.326} \\
& 720  & 0.399 & 0.394 & \textbf{0.364} & \textbf{0.378} & 0.365 & 0.379 & \textbf{0.364} & \textbf{0.379} & \textbf{0.389} & \textbf{0.399} & 0.397 & 0.409 & 0.361 & 0.381 & \textbf{0.355} & \textbf{0.379} & 0.375 & 0.393 & \textbf{0.371} & \textbf{0.391} \\
\cline{2-22}
& \textbf{Avg}& 0.283 & 0.322 & \textbf{0.256} & \textbf{0.308} & \textbf{0.255} & 0.309 & \textbf{0.255} & \textbf{0.308} & 0.281 & 0.335 & \textbf{0.273} & \textbf{0.328} & 0.258 & 0.314 & \textbf{0.256} & \textbf{0.309} & 0.260 & 0.314 & \textbf{0.259} & \textbf{0.313} \\

\midrule

\multirow{5}{*}{\rotatebox[origin=c]{90}{ETTh1}} 
& 96  & 0.379 & 0.401 & \textbf{0.361} & \textbf{0.389} & 0.359 & 0.383 & \textbf{0.358} & \textbf{0.381} & 0.409 & 0.415 & \textbf{0.387} & \textbf{0.406} & 0.383 & 0.403 & \textbf{0.359} & \textbf{0.386} & 0.365 & 0.383 & \textbf{0.365} & \textbf{0.383} \\
& 192  & 0.423 & 0.430 & \textbf{0.410} & \textbf{0.418} & 0.398 & 0.406 & \textbf{0.398} & \textbf{0.406} & 0.507 & 0.462 & \textbf{0.429} & \textbf{0.436} & 0.409 & 0.424 & \textbf{0.397} & \textbf{0.412} & 0.417 & 0.426 & \textbf{0.405} & \textbf{0.409} \\
& 336  & 0.440 & 0.443 & \textbf{0.428} & \textbf{0.431} & \textbf{0.426} & \textbf{0.422} & 0.428 & 0.423 & 0.459 & \textbf{0.451} & \textbf{0.457} & 0.457 & 0.440 & 0.446 & \textbf{0.422} & \textbf{0.433} & 0.443 & 0.436 & \textbf{0.440} & \textbf{0.431} \\
& 720  & 0.455 & 0.475 & \textbf{0.440} & \textbf{0.460} & \textbf{0.430} & \textbf{0.450} & 0.433 & 0.452 & 0.526 & \textbf{0.506} & \textbf{0.523} & 0.519 & 0.481 & 0.482 & \textbf{0.475} & \textbf{0.479} & \textbf{0.456} & \textbf{0.473} & 0.467 & 0.481 \\
\cline{2-22}
& \textbf{Avg}& 0.424 & 0.437 & \textbf{0.410} & \textbf{0.425} & \textbf{0.403} & \textbf{0.415} & 0.404 & 0.416 & 0.475 & 0.459 & \textbf{0.449} & \textbf{0.455} & 0.428 & 0.439 & \textbf{0.413} & \textbf{0.428} & 0.420 & 0.430 & \textbf{0.419} & \textbf{0.426} \\

\midrule
\multirow{5}{*}{\rotatebox[origin=c]{90}{ETTh2}} 
& 96  & 0.295 & 0.342 & \textbf{0.277} & \textbf{0.334} & 0.273 & 0.331 & \textbf{0.272} & \textbf{0.330} & 0.343 & 0.373 & \textbf{0.292} & \textbf{0.348} & 0.292 & 0.346 & \textbf{0.288} & \textbf{0.338} & 0.278 & 0.333 & \textbf{0.278} & \textbf{0.333} \\
& 192  & 0.361 & 0.388 & \textbf{0.347} & \textbf{0.379} & 0.347 & 0.382 & \textbf{0.338} & \textbf{0.375} & 0.410 & 0.413 & \textbf{0.370} & \textbf{0.399} & 0.354 & 0.385 & \textbf{0.350} & \textbf{0.378} & \textbf{0.350} & \textbf{0.382} & 0.352 & 0.385 \\
& 336  & 0.426 & 0.428 & \textbf{0.372} & \textbf{0.405} & 0.394 & 0.415 & \textbf{0.371} & \textbf{0.402} & \textbf{0.419} & 0.434 & 0.421 & \textbf{0.431} & 0.383 & 0.407 & \textbf{0.365} & \textbf{0.400} & 0.403 & 0.424 & \textbf{0.389} & \textbf{0.413} \\
& 720  & 0.427 & 0.446 & \textbf{0.408} & \textbf{0.432} & 0.405 & 0.433 & \textbf{0.400} & \textbf{0.430} & \textbf{0.448} & \textbf{0.461} & 0.497 & 0.498 & 0.587 & 0.549 & \textbf{0.387} & \textbf{0.420} & 0.556 & 0.518 & \textbf{0.546} & \textbf{0.515} \\
\cline{2-22}
& \textbf{Avg}& 0.377 & 0.401 & \textbf{0.351} & \textbf{0.388} & 0.355 & 0.390 & \textbf{0.345} & \textbf{0.384} & 0.405 & 0.420 & \textbf{0.395} & \textbf{0.419} & 0.404 & 0.422 & \textbf{0.348} & \textbf{0.384} & 0.397 & 0.414 & \textbf{0.391} & \textbf{0.412} \\

\midrule

\multirow{5}{*}{\rotatebox[origin=c]{90}{Exchange}} 
& 96  & 0.094 & 0.216 & \textbf{0.090} & \textbf{0.212} & 0.092 & 0.213 & \textbf{0.088} & \textbf{0.206} & 0.109 & 0.239 & \textbf{0.107} & \textbf{0.237} & 0.098 & 0.221 & \textbf{0.090} & \textbf{0.210} & \textbf{0.082} & \textbf{0.201} & 0.095 & 0.214 \\
& 192  & 0.190 & 0.313 & \textbf{0.183} & \textbf{0.308} & 0.189 & 0.307 & \textbf{0.184} & \textbf{0.302} & \textbf{0.266} & 0.377 & 0.269 & \textbf{0.367} & 0.210 & 0.329 & \textbf{0.195} & \textbf{0.317} & 0.174 & 0.299 & \textbf{0.154} & \textbf{0.283} \\
& 336  & \textbf{0.331} & \textbf{0.417} & 0.342 & 0.424 & 0.392 & 0.456 & \textbf{0.354} & \textbf{0.428} & \textbf{0.396} & \textbf{0.467} & 0.579 & 0.566 & 0.468 & 0.506 & \textbf{0.379} & \textbf{0.443} & 0.347 & 0.427 & \textbf{0.276} & \textbf{0.392} \\
& 720  & 1.138 & 0.784 & \textbf{0.792} & \textbf{0.659} & 0.901 & 0.714 & \textbf{0.863} & \textbf{0.697} & 0.841 & 0.691 & \textbf{0.791} & \textbf{0.678} & 1.082 & 0.793 & \textbf{0.791} & \textbf{0.661} & 1.493 & 0.893 & \textbf{0.961} & \textbf{0.737} \\
\cline{2-22}
& \textbf{Avg}& 0.438 & 0.433 & \textbf{0.352} & \textbf{0.401} & 0.394 & 0.423 & \textbf{0.372} & \textbf{0.408} & \textbf{0.403} & \textbf{0.444} & 0.437 & 0.462 & 0.465 & 0.462 & \textbf{0.364} & \textbf{0.408} & 0.524 & 0.455 & \textbf{0.372} & \textbf{0.407} \\

\midrule
\multirow{5}{*}{\rotatebox[origin=c]{90}{Traffic}} 
& 96  & 0.406 & 0.285 & \textbf{0.398} & \textbf{0.275} & \textbf{0.424} & \textbf{0.288} & 0.425 & 0.289 & 0.398 & 0.275 & \textbf{0.389} & \textbf{0.267} & 0.402 & 0.264 & \textbf{0.399} & \textbf{0.257} & \textbf{0.430} & 0.269 & 0.434 & \textbf{0.268} \\
& 192  & \textbf{0.421} & \textbf{0.289} & 0.425 & 0.290 & \textbf{0.433} & 0.296 & 0.434 & \textbf{0.296} & 0.417 & 0.286 & \textbf{0.407} & \textbf{0.277} & 0.416 & 0.271 & \textbf{0.416} & \textbf{0.265} & \textbf{0.435} & 0.302 & 0.444 & \textbf{0.272} \\
& 336  & \textbf{0.435} & 0.296 & 0.437 & \textbf{0.296} & \textbf{0.445} & 0.304 & 0.446 & \textbf{0.303} & 0.431 & 0.294 & \textbf{0.422} & \textbf{0.282} & 0.429 & 0.276 & \textbf{0.409} & \textbf{0.265} & \textbf{0.447} & 0.308 & 0.453 &  \textbf{0.277} \\
& 720  & \textbf{0.465} & \textbf{0.309} & 0.472 & 0.316 & 0.474 & 0.322 & \textbf{0.474} & \textbf{0.320} & 0.463 & 0.312 & \textbf{0.456} & \textbf{0.301} & 0.456 & 0.293 & \textbf{0.447} & \textbf{0.281} & \textbf{0.476} & 0.327 & 0.477 & \textbf{0.293} \\
\cline{2-22}
& \textbf{Avg}& \textbf{0.432} & 0.295 & 0.433 & \textbf{0.294} & \textbf{0.444} & 0.303 & 0.445 & \textbf{0.302} & 0.427 & 0.292 & \textbf{0.419} & \textbf{0.282} & 0.426 & 0.276 & \textbf{0.418} & \textbf{0.267} & \textbf{0.447} & 0.302 & 0.452 & \textbf{0.278} \\

\midrule

\multirow{5}{*}{\rotatebox[origin=c]{90}{Weather}} 
& 96  & 0.148 & 0.188 & \textbf{0.147} & \textbf{0.187} & 0.176 & 0.214 & \textbf{0.172} & \textbf{0.211} & 0.162 & 0.204 & \textbf{0.152} & \textbf{0.191} & 0.155 & 0.189 & \textbf{0.151} & \textbf{0.186} & 0.178 & 0.215 & \textbf{0.178} & \textbf{0.213} \\
& 192  & 0.193 & 0.232 & \textbf{0.189} & \textbf{0.227} & 0.217 & 0.251 & \textbf{0.217} & \textbf{0.251} & 0.213 & 0.249 & \textbf{0.195} & \textbf{0.238} & \textbf{0.194} & 0.230 & 0.195 & \textbf{0.230} & 0.218 & 0.252 & \textbf{0.217} & \textbf{0.252} \\
& 336  & 0.247 & 0.274 & \textbf{0.242} & \textbf{0.270} & 0.265 & 0.288 & \textbf{0.263} & \textbf{0.286} & 0.263 & 0.287 & \textbf{0.244} & \textbf{0.279} & 0.256 & 0.279 & \textbf{0.246} & \textbf{0.271} & 0.261 & \textbf{0.291} & \textbf{0.261} & 0.292 \\
& 720  & \textbf{0.318} & \textbf{0.323} & 0.325 & 0.327 & 0.333 & 0.334 & \textbf{0.332} & \textbf{0.334} & 0.339 & 0.340 & \textbf{0.315} & \textbf{0.338} & 0.327 & 0.327 & \textbf{0.327} & \textbf{0.326} & 0.326 & 0.354 & \textbf{0.322} & \textbf{0.344} \\
\cline{2-22}
& \textbf{Avg}& 0.227 & 0.254 & \textbf{0.226} & \textbf{0.253} & 0.248 & 0.272 & \textbf{0.246} & \textbf{0.271} & 0.244 & 0.270 & \textbf{0.227} & \textbf{0.262} & 0.233 & 0.256 & \textbf{0.230} & \textbf{0.253} & 0.246 & 0.278 & \textbf{0.245} & \textbf{0.275} \\

\midrule
\multirow{5}{*}{\rotatebox[origin=c]{90}{Solar-Energy}} 
& 96  & 0.234 & \textbf{0.263} & \textbf{0.222} & 0.264 & 0.224 & 0.222 & \textbf{0.220} & \textbf{0.222} & 0.188 & 0.210 & \textbf{0.174} & \textbf{0.207} & 0.204 & 0.229 & \textbf{0.186} & \textbf{0.217} & \textbf{0.223} & 0.293 & 0.234 & \textbf{0.214} \\
& 192  & 0.266 & 0.287 & \textbf{0.251} & \textbf{0.278} & 0.256 & 0.242 & \textbf{0.248} & \textbf{0.239} & 0.213 & \textbf{0.226} & \textbf{0.204} & 0.233 & 0.211 & 0.225 & \textbf{0.201} & \textbf{0.224} & \textbf{0.251} & 0.311  & 0.266 & \textbf{0.234} \\
& 336  & 0.289 & 0.304 & \textbf{0.240} & \textbf{0.263} & 0.279 & 0.257 & \textbf{0.272} & \textbf{0.254} & 0.228 & 0.235 & \textbf{0.200} & \textbf{0.233} & 0.217 & \textbf{0.236} & \textbf{0.212} & 0.238 & \textbf{0.272} & 0.327  & 0.289 & \textbf{0.248} \\
& 720  & 0.290 & 0.315 & \textbf{0.254} & \textbf{0.264} & 0.280 & 0.260 & \textbf{0.261} & \textbf{0.250} & 0.233 & 0.237 & \textbf{0.204} & \textbf{0.227} & \textbf{0.220} & \textbf{0.236} & 0.226 & 0.237 & \textbf{0.274} & 0.330 & 0.288 & \textbf{0.248} \\
\cline{2-22}
& \textbf{Avg}& 0.270 & 0.292 & \textbf{0.242} & \textbf{0.267} & 0.260 & 0.245 & \textbf{0.250} & \textbf{0.241} & 0.216 & 0.227 & \textbf{0.196} & \textbf{0.225} & 0.213 & 0.232 & \textbf{0.206} & \textbf{0.229} & \textbf{0.255} & 0.315 & 0.269 & \textbf{0.236} \\

\midrule

\multirow{5}{*}{\rotatebox[origin=c]{90}{Wind}} 
& 96  & 0.750 & 0.660 & \textbf{0.734} & \textbf{0.651} & 0.720 & 0.644 & \textbf{0.720} & \textbf{0.644} & 0.741 & 0.664 & \textbf{0.703} & \textbf{0.640} & 0.733 & 0.651 & \textbf{0.720} & \textbf{0.646} & \textbf{0.709} & 0.646 & 0.710 & \textbf{0.646} \\
& 192  & 0.770 & 0.681 & \textbf{0.759} & \textbf{0.669} & 0.759 & 0.669 & \textbf{0.759} & \textbf{0.669} & 0.779 & 0.685 & \textbf{0.735} & \textbf{0.663} & 0.758 & 0.677 & \textbf{0.758} & \textbf{0.671} & \textbf{0.743} & 0.672 & 0.744 & \textbf{0.672} \\
& 336  & 0.793 & 0.695 & \textbf{0.780} & \textbf{0.683} & 0.780 & 0.683 & \textbf{0.780} & \textbf{0.683} & 0.789 & 0.696 & \textbf{0.744} & \textbf{0.672} & 0.800 & 0.692 & \textbf{0.771} & \textbf{0.681} & 0.762 & 0.685 & \textbf{0.762} & \textbf{0.685} \\
& 720  & 0.818 & 0.711 & \textbf{0.804} & \textbf{0.703} & 0.811 & 0.704 & \textbf{0.810} & \textbf{0.704} & 0.812 & 0.712 & \textbf{0.745} & \textbf{0.679} & 0.811 & 0.706 & \textbf{0.794} & \textbf{0.700} & \textbf{0.781} & 0.703 & 0.782 & \textbf{0.703} \\
\cline{2-22}
& \textbf{Avg}& 0.783 & 0.687 & \textbf{0.769} & \textbf{0.677} & 0.768 & \textbf{0.675} & \textbf{0.767} & \textbf{0.675} & 0.780 & 0.689 & \textbf{0.732} & \textbf{0.664} & 0.776 & 0.682 & \textbf{0.761} & \textbf{0.675} & \textbf{0.749} & \textbf{0.677} & 0.750 & \textbf{0.677} \\

\bottomrule
\end{tabular}
}
\end{table}

\paragraph{Full Hyperparameter Search.}
\label{appendix:hparam_results}
Table~\ref{tab::long-term-search} compares MSAR with five baselines after tuning all methods within a shared search space. The look-back window is selected from $\{96, 192, 320, 512\}$, with common candidate ranges for model width, depth, and learning rate; the full search protocol is provided in Appendix~\ref{sec:hparam_search}. We report MSE and MAE at four prediction lengths on the four ETT datasets and Weather. MSAR achieves the lowest average MSE and MAE on ETTm1, ETTm2, and ETTh1. On ETTh2, it achieves the lowest average MAE, while DUET obtains a lower average MSE. On Weather, MSAR ties SimpleTM in average MAE at the reported precision, while SimpleTM achieves a lower average MSE. These results show that MSAR remains competitive when the baselines are tuned beyond the fixed-setting comparison.

\begin{table*}[!t]
    \caption{Full hyperparameter search results on five datasets with prediction lengths $\{96, 192, 336, 720\}$. Each model is evaluated with its best configuration within the shared search space; the look-back window is selected from $\{96, 192, 320, 512\}$. Avg. denotes the average across prediction lengths. Search details are provided in Appendix~\ref{sec:hparam_search}.}
	\label{tab::long-term-search}
	\setlength{\tabcolsep}{3pt}
    \footnotesize
	\centering
	\begin{threeparttable}
		\begin{tabular}{c|c|c c|c c|c c|c c|c c|c c}

			\toprule
			\multicolumn{2}{c}{\multirow{2}{*}{\scalebox{1.1}{Models}}}& \multicolumn{2}{c}{MSAR} & \multicolumn{2}{c}{SimpleTM} & \multicolumn{2}{c}{DUET} & \multicolumn{2}{c}{iTransformer} & \multicolumn{2}{c}{PatchTST} & \multicolumn{2}{c}{DLinear}\\

			\multicolumn{2}{c}{} & \multicolumn{2}{c}{\scalebox{0.8}{\textbf{Ours}}} &
            \multicolumn{2}{c}{\scalebox{0.8}{\citeyearpar{chen2025simpletm}}} & 
            \multicolumn{2}{c}{\scalebox{0.8}{\citeyearpar{qiu2025duet}}} &
            \multicolumn{2}{c}{\scalebox{0.8}{\citeyearpar{liu2024itransformer}}} & 
            \multicolumn{2}{c}{\scalebox{0.8}{\citeyearpar{nie2022time}}} & 
            \multicolumn{2}{c}{\scalebox{0.8}{\citeyearpar{zeng2023transformers}}}\\

			\cmidrule(lr){3-4} \cmidrule(lr){5-6} \cmidrule(lr){7-8} \cmidrule(lr){9-10} \cmidrule(lr){11-12} \cmidrule(lr){13-14}
			\multicolumn{2}{c}{Metric}& MSE & MAE & MSE & MAE & MSE & MAE & MSE & MAE & MSE & MAE & MSE & MAE\\

			\toprule
                \multirow{5}{*}{\rotatebox[origin=c]{90}{ETTm1}} 
                & 96  & \secondres{0.289} & \boldres{0.335} & \boldres{0.280} & \secondres{0.337} & 0.300 & 0.345 & 0.299 & 0.348 & \secondres{0.289} & 0.343 & 0.299 & 0.343 \\

                & 192 & \boldres{0.325} & \boldres{0.355} & \secondres{0.326} & \secondres{0.365} & 0.335 & 0.366 & 0.334 & 0.373 & 0.332 & 0.367 & 0.336 & 0.367 \\

                & 336 & \boldres{0.355} & \boldres{0.376} & \secondres{0.362} & 0.387 & 0.364 & \secondres{0.383} & 0.370 & 0.394 & \secondres{0.362} & 0.389 & 0.367 & 0.385 \\

                & 720 & \secondres{0.410} & \boldres{0.406} & 0.417 & 0.422 & 0.417 & \secondres{0.413} & 0.422 & 0.426 & \boldres{0.408} & 0.420 & 0.420 & 0.418 \\

                \cmidrule(lr){2-14}
                & \emph{Avg.} & \boldres{0.345} & \boldres{0.368} & \secondres{0.346} & 0.378 & 0.354 & \secondres{0.377} & 0.356 & 0.385 & 0.348 & 0.380 & 0.356 & 0.378 \\

			\midrule
			\multirow{5}{*}{\rotatebox[origin=c]{90}{ETTm2}} 
                & 96  & \boldres{0.160} & \boldres{0.246} & 0.168 & \secondres{0.252} & \secondres{0.162} & 0.253 & 0.180 & 0.263 & 0.165 & 0.257 & 0.165 & 0.260 \\

            & 192 & \boldres{0.216} & \boldres{0.284} & 0.223 & 0.297 & \secondres{0.217} & \secondres{0.290} & 0.236 & 0.307 & 0.223 & 0.294 & 0.219 & 0.297 \\

            & 336 & \boldres{0.269} & \boldres{0.321} & 0.273 & 0.328 & \secondres{0.270} & \secondres{0.325} & 0.287 & 0.340 & 0.272 & 0.333 & 0.290 & 0.345 \\

            & 720 & \boldres{0.359} & \boldres{0.377} & 0.364 & 0.386 & \secondres{0.361} & \secondres{0.384} & 0.364 & 0.391 & 0.363 & 0.386 & 0.375 & 0.403 \\

                \cmidrule(lr){2-14}
                & \emph{Avg.} & \boldres{0.251} & \boldres{0.307} & 0.257 & 0.316 & \secondres{0.253} & \secondres{0.313} & 0.267 & 0.325 & 0.256 & 0.318 & 0.262 & 0.326 \\
            
			\midrule
			\multirow{5}{*}{\rotatebox[origin=c]{90}{ETTh1}} 
			& 96  & \boldres{0.356} & \boldres{0.381} & 0.375 & 0.394 & \secondres{0.364} & \secondres{0.392} & 0.380 & 0.401 & 0.370 & 0.397 & 0.368 & 0.393 \\

            & 192 & \boldres{0.395} & \boldres{0.407} & 0.412 & 0.429 & \secondres{0.397} & \secondres{0.413} & 0.423 & 0.431 & 0.404 & 0.423 & 0.400 & 0.417 \\
            
            & 336 & 0.426 & \boldres{0.424} & 0.429 & 0.443 & \boldres{0.422} & \secondres{0.432} & 0.437 & 0.450 & \secondres{0.423} & 0.437 & 0.430 & 0.441 \\
            
            & 720 & \boldres{0.428} & \boldres{0.449} & 0.447 & 0.472 & \secondres{0.443} & \secondres{0.463} & 0.460 & 0.471 & 0.444 & 0.464 & 0.477 & 0.497 \\

                \cmidrule(lr){2-14}
                & \emph{Avg.} & \boldres{0.401} & \boldres{0.417} & 0.416 & 0.434 & \secondres{0.406} & \secondres{0.425} & 0.425 & 0.438 & 0.410 & 0.430 & 0.419 & 0.437 \\
            
			\midrule
			\multirow{5}{*}{\rotatebox[origin=c]{90}{ETTh2}} 
            
			& 96  & \secondres{0.267} & \boldres{0.328} & 0.291 & 0.339 & \boldres{0.265} & \secondres{0.334} & 0.294 & 0.345 & 0.290 & 0.342 & 0.286 & 0.351 \\

            & 192 & \secondres{0.331} & \boldres{0.362} & 0.349 & 0.390 & \boldres{0.324} & \secondres{0.373} & 0.358 & 0.395 & 0.354 & 0.397 & 0.352 & 0.394 \\

            & 336 & \secondres{0.364} & \boldres{0.399} & 0.381 & 0.413 & \boldres{0.352} & \secondres{0.399} & 0.385 & 0.415 & 0.384 & 0.413 & 0.439 & 0.456 \\

            & 720 & \secondres{0.397} & \boldres{0.430} & 0.410 & 0.442 & \boldres{0.393} & \secondres{0.432} & 0.412 & 0.436 & 0.408 & 0.442 & 0.570 & 0.530 \\

			\cmidrule(lr){2-14}
			& \emph{Avg.}  & \secondres{0.340} & \boldres{0.380} & 0.358 & 0.396 & \boldres{0.334} & \secondres{0.384} & 0.362 & 0.398 & 0.359 & 0.399 & 0.412 & 0.433 \\
            
			\midrule
			\multirow{5}{*}{\rotatebox[origin=c]{90}{Weather}}
            
			& 96  & \secondres{0.145} & \boldres{0.189} & \boldres{0.143} & 0.195 & 0.168 & 0.221 & 0.155 & 0.206 & 0.145 & \secondres{0.194} & 0.170 & 0.229 \\

        & 192 & \boldres{0.188} & \boldres{0.230} & \secondres{0.188} & \secondres{0.236} & 0.212 & 0.258 & 0.199 & 0.249 & 0.191 & 0.237 & 0.212 & 0.268 \\

        & 336 & 0.257 & \boldres{0.284} & \boldres{0.238} & \secondres{0.276} & 0.258 & 0.292 & 0.247 & 0.284 & \secondres{0.243} & 0.279 & 0.258 & 0.307 \\

       & 720 & 0.323 & 0.332 & \boldres{0.312} & \boldres{0.327} & 0.324 & 0.338 & 0.318 & 0.336 & \secondres{0.313} & \secondres{0.330} & 0.321 & 0.359 \\

                \cmidrule(lr){2-14}
			& \emph{Avg.} & \secondres{0.228} & \boldres{0.259} & \boldres{0.220} & \secondres{0.259} & 0.240 & 0.277 & 0.230 & 0.269 & \secondres{0.223} & 0.260 & 0.240 & 0.291 \\   

            \midrule
            \multicolumn{2}{c}{$1^{\text{st}}$ \emph{Count}} & \multicolumn{2}{c}{\boldres{37}}& \multicolumn{2}{c}{\secondres{6}} & \multicolumn{2}{c}{\secondres{6}} & \multicolumn{2}{c}{0} & \multicolumn{2}{c}{1} & \multicolumn{2}{c}{0}\\
			\toprule
		\end{tabular}
	\end{threeparttable}
\end{table*}

\paragraph{Short-term forecasting results.}  
Table~\ref{tab:tabshort} reports short-term forecasting results on the PEMS datasets with a prediction length of $12$. 
We evaluate three common metrics—MAE, MAPE, and RMSE—to provide a comprehensive view of performance. 
MSAR lowers MAE for SimpleTM, iTransformer, PatchTST, and DLinear on all four subsets. DUET benefits less consistently, with small MAE and RMSE increases on PEMS04 and PEMS08. Thus, the gains extend to short-term forecasting, but their size and consistency depend on the backbone.

\begin{table}
\caption{
Short-term forecasting results on PEMS datasets (prediction length = 12). 
Lower MAE, MAPE, and RMSE indicate better performance.}
\label{tab:tabshort}
\centering
\renewcommand{\arraystretch}{1.15}
\setlength{\tabcolsep}{4pt}
\newcolumntype{C}[1]{>{\centering\arraybackslash}p{#1}}
\resizebox{0.9\textwidth}{!}{\begin{tabular}{C{1.5cm}C{1.2cm}C{0.95cm}C{1.0cm}C{0.95cm}C{1.0cm}C{0.95cm}C{1.0cm}C{0.95cm}C{1.0cm}C{0.95cm}C{1.0cm}}\toprule
\multirow{2}{*}{Dataset} & \multirow{2}{*}{Metric} 
& \multicolumn{2}{c}{SimpleTM} 
& \multicolumn{2}{c}{DUET} 
& \multicolumn{2}{c}{iTransformer} 
& \multicolumn{2}{c}{PatchTST} 
& \multicolumn{2}{c}{DLinear} \\
\cmidrule(lr){3-4} \cmidrule(lr){5-6} \cmidrule(lr){7-8} \cmidrule(lr){9-10} \cmidrule(lr){11-12}
& & Base & +Ours & Base & +Ours & Base & +Ours & Base & +Ours & Base & +Ours \\\midrule

\multirow{3}{*}{PEMS03} & MAE  & 19.147 & \textbf{15.627} & 18.853 & \textbf{18.402} & 15.870 & \textbf{15.308} & 16.974 & \textbf{15.209} & 17.411 & \textbf{16.743} \\& MAPE  & 19.527 & \textbf{15.796} & 22.031 & \textbf{20.698} & 15.945 & \textbf{15.459} & 19.376 & \textbf{15.905} & 15.876 & \textbf{15.509} \\& RMSE  & 30.307 & \textbf{24.933} & 29.607 & \textbf{29.240} & 25.563 & \textbf{24.648} & 26.517 & \textbf{24.310} & 29.619 & \textbf{27.909} \\\midrule
\multirow{3}{*}{PEMS04} & MAE  & 27.192 & \textbf{21.651} & \textbf{24.581} & 24.693 & 22.033 & \textbf{20.617} & 23.005 & \textbf{21.022} & 24.359 & \textbf{22.546} \\& MAPE  & 18.064 & \textbf{13.462} & 18.659 & \textbf{18.202} & 14.037 & \textbf{13.019} & 15.779 & \textbf{13.846} & 14.616 & \textbf{13.538} \\& RMSE  & 41.719 & \textbf{34.425} & \textbf{37.486} & 37.614 & 34.834 & \textbf{33.023} & 35.616 & \textbf{33.300} & 38.966 & \textbf{36.244} \\\midrule
\multirow{3}{*}{PEMS07} & MAE  & 26.999 & \textbf{22.770} & 26.862 & \textbf{26.577} & 23.064 & \textbf{21.863} & 24.010 & \textbf{21.911} & 25.519 & \textbf{24.655} \\& MAPE  & 12.158 & \textbf{9.675} & 13.401 & \textbf{13.258} & 10.119 & \textbf{9.544} & 10.821 & \textbf{9.311} & 10.869 & \textbf{10.384} \\& RMSE  & 41.257 & \textbf{35.833} & 40.314 & \textbf{40.114} & 36.330 & \textbf{34.762} & 36.711 & \textbf{33.92} & 40.313 & \textbf{39.109} \\\midrule
\multirow{3}{*}{PEMS08} & MAE  & 22.261 & \textbf{16.412} & 19.922 & \textbf{19.976} & 17.350 & \textbf{15.851} & 18.591 & \textbf{16.057} & 20.353 & \textbf{18.090} \\& MAPE  & 14.124 & \textbf{10.086} & 13.631 & \textbf{13.459} & 10.963 & \textbf{9.990} & 12.569 & \textbf{10.419} & 12.376 & \textbf{11.047} \\& RMSE  & 34.455 & \textbf{26.059} & 30.199 & \textbf{30.720} & 27.459 & \textbf{25.312} & 28.172 & \textbf{25.267} & 32.658 & \textbf{29.043} \\

\bottomrule
\end{tabular}
}
\end{table}

\section{Efficiency Study}
\label{appendix:efficiency}
To assess the inference efficiency of our MSAR framework, we report the wall-clock runtime of four backbone architectures—DLinear~\citep{zeng2023transformers}, TimesNet~\citep{wu2022timesnet}, PatchTST~\citep{nie2022time}, and iTransformer~\citep{liu2024itransformer}—both with and without MSAR integration. Additionally, we compare against SegRNN~\citep{lin2023segrnn}, a representative autoregressive baseline that sequentially generates future time steps.

\begin{figure}[htbp]
    \centering
    \begin{minipage}[c]{0.49\linewidth} \centering
        \includegraphics[width=\linewidth]{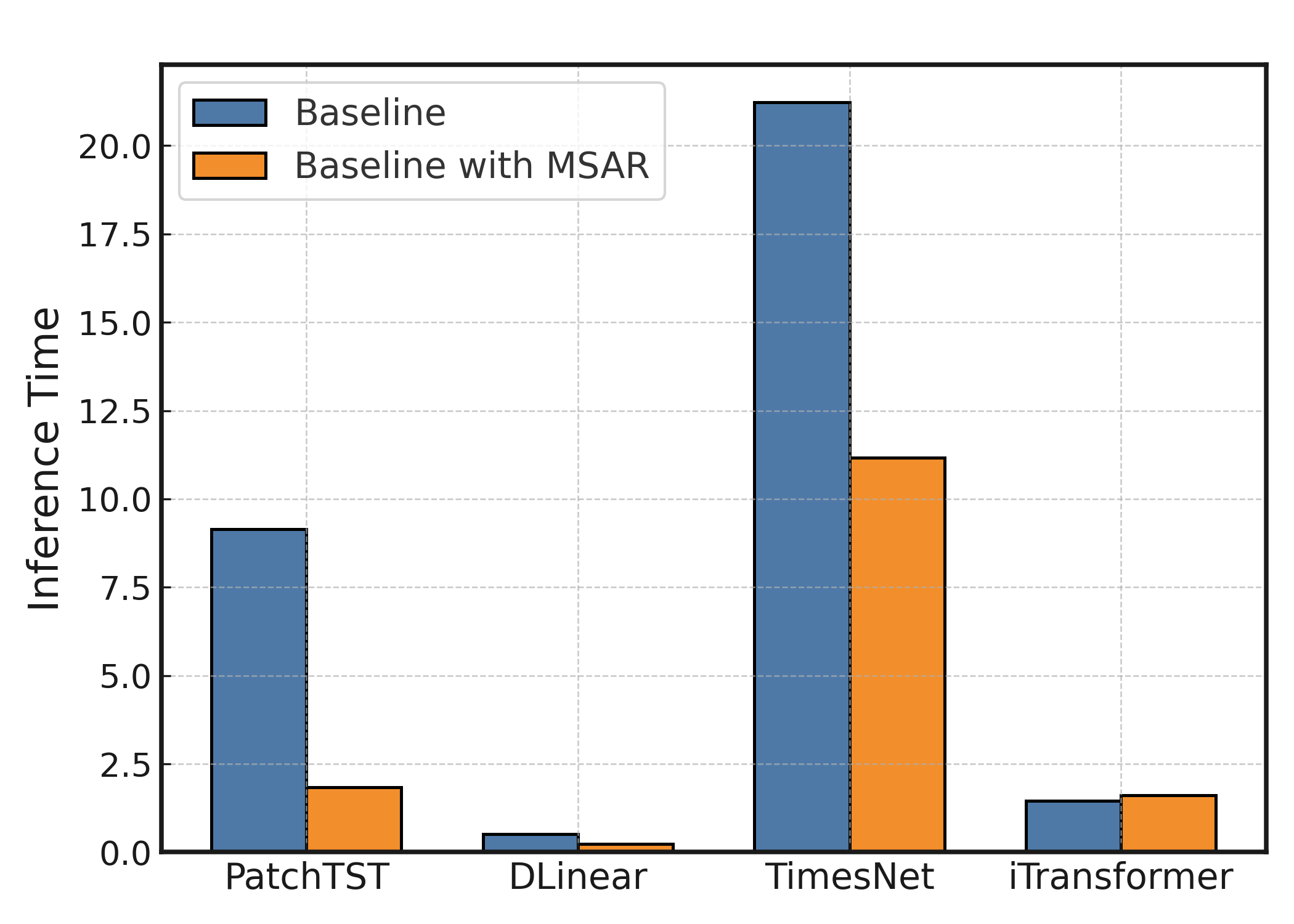}
        \caption{Inference efficiency of baseline models vs. MSAR-enhanced models (batch size 32, input=960, pred=336).}
        \label{fig:efficiency}
    \end{minipage}
    \hfill
    \begin{minipage}[c]{0.49\linewidth}
    \centering
    \renewcommand{\arraystretch}{1.1}
    \rowcolors{3}{gray!10}{white}
    \begin{tabular}{@{}lcc@{}}
        \toprule
        \textbf{Model} & \textbf{H} & \textbf{Inference Time (s)} \\
        \midrule
        SegRNN$^{6}$ & 336 & 7.23 \\
        SegRNN$^{12}$ & 336 & 4.06 \\
        SegRNN$^{6}$ & 720 & 8.53 \\
        SegRNN$^{12}$ & 720 & 5.01 \\
        \midrule
        PatchTST$^*$ & 336 & 2.13 \\
        PatchTST$^*$ & 720 & 2.30 \\
        iTransformer$^*$ & 336 & 1.84 \\
        iTransformer$^*$ & 720 & 1.94 \\
        DLinear$^*$ & 336 & 0.31 \\
        DLinear$^*$ & 720 & 0.36 \\
        \bottomrule
    \end{tabular}
    \captionof{table}{Inference time (in seconds) for different models and prediction lengths \textbf{H}. 
    $^*$ indicates MSAR-enhanced models. SegRNN$^{6}$ and SegRNN$^{12}$ denote segment length.}
    \label{tab:efficiency}
\end{minipage}
\end{figure}

Figure~\ref{fig:efficiency} compares the inference time of baseline models with their MSAR-enhanced counterparts under identical input-output settings (input length = 960, prediction length = 336, batch size = 32). For most backbones, including TimesNet~\citep{wu2022timesnet} and PatchTST~\citep{nie2022time}, the hierarchical design of MSAR improves both prediction accuracy and runtime by reducing redundant computation through coarse-scale anchoring. However, we observe an increase in inference time when integrating MSAR with iTransformer~\citep{liu2024itransformer}. This is expected, as iTransformer encodes inputs via a global linear projection into the latent space without explicit dependency on sequence length. The hierarchical multi-scale processing in MSAR introduces additional overhead in such architectures, where lookback windows do not otherwise impact runtime.

Table~\ref{tab:efficiency} collectively demonstrates the inference efficiency of MSAR-enhanced models compared to autoregressive baselines on the ETTh1 dataset. It shows that MSAR consistently achieves significant inference speedups over SegRNN~\citep{lin2023segrnn}, as it avoids full-step autoregression by leveraging parallel decoding within each scale. MSAR achieves favorable efficiency-accuracy trade-offs across backbones, demonstrating its generalizability and practical utility for scalable time series forecasting.

\section{Additional Prediction Results}

In this section, we provide additional qualitative results on the ETTh1 dataset.
All experiments are conducted with a single-layer encoder (\texttt{Layer=1}),
two attention heads (\texttt{Head=2}), and an interval configuration of
\texttt{interval\_list = [3, 1]}. We evaluate the model under four input sequence length settings: \texttt{[32, 96]}, \texttt{[32, 192]}, \texttt{[32, 384]}, and \texttt{[32, 512]}. The visualizations below are generated using our proposed \textbf{DUET+MSAR} model, showing its prediction accuracy compared to the ground-truth time series.

\begin{figure}[ht]
    \centering
    \includegraphics[width=\linewidth]{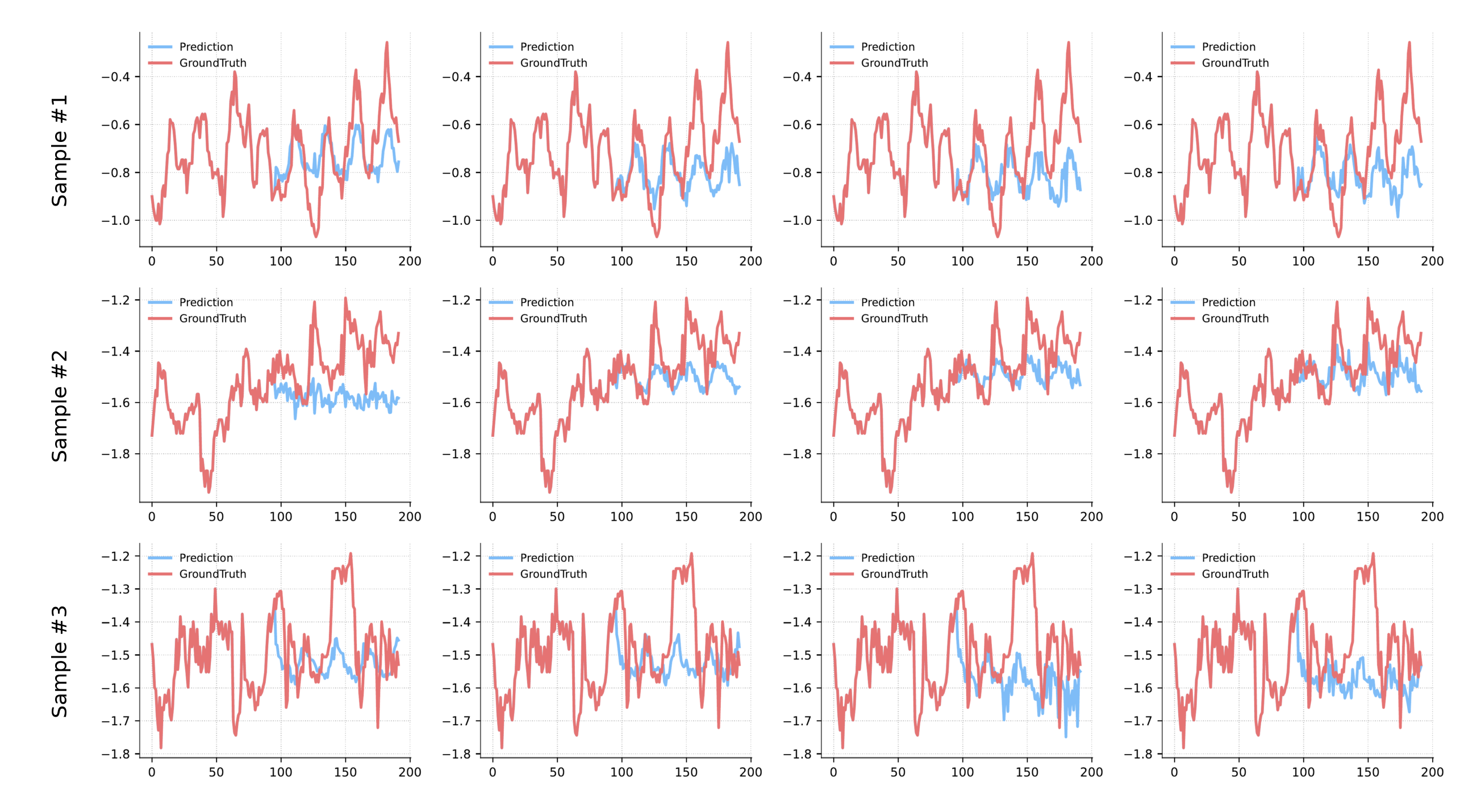}
    \caption{Visualization of model predictions and ground truth on the ETTh1 dataset.
        Results are produced by the \textbf{DUET+MSAR} model.}
    \label{fig:appendix_pred_gt}
\end{figure}

The visualizations in Figure~\ref{fig:appendix_pred_gt} are generated using our proposed \textbf{DUET+MSAR}
model, illustrating its prediction accuracy compared to the ground-truth
time series.

\clearpage

\section*{NeurIPS Paper Checklist}

\begin{enumerate}

\item {\bf Claims}
    \item[] Question: Do the main claims made in the abstract and introduction accurately reflect the paper's contributions and scope?
    \item[] Answer: \answerYes{} \item[] Justification: The introduction clearly states the three key contributions of MSAR: scale-wise aligned modeling, scale-wise autoregression, and a modular model-agnostic design. These claims are directly supported by the experimental results in Section~\ref{sec:Forecasting_Results} and the ablation study in Section~\ref{sec:ablation}.

\item {\bf Limitations}
    \item[] Question: Does the paper discuss the limitations of the work performed by the authors?
    \item[] Answer: \answerYes{} \item[] Justification: The ``Limitations and Future Work'' paragraph at the end of the Experiments section discusses coarse-scale error propagation, additional training stages, and the lack of improvements in some backbone--dataset settings. It also acknowledges that variability across random seeds is not systematically characterized and that real-time deployment may require adaptive scheduling or scale selection.

\item {\bf Theory assumptions and proofs}
    \item[] Question: For each theoretical result, does the paper provide the full set of assumptions and a complete (and correct) proof?
    \item[] Answer: \answerNA{} \item[] Justification: The paper presents an algorithmic framework and empirical analysis without formal theoretical results requiring mathematical proofs.

    \item {\bf Experimental result reproducibility}
    \item[] Question: Does the paper fully disclose all the information needed to reproduce the main experimental results of the paper to the extent that it affects the main claims and/or conclusions of the paper (regardless of whether the code and data are provided or not)?
    \item[] Answer: \answerYes{} \item[] Justification: The paper provides full experimental details in Appendix~\ref{appendix:exp_details}, including dataset splits, hyperparameter search spaces, optimizer (Adam, lr=$10^{-3}$), training settings, and the MSAR-specific configuration (\texttt{interval\_list}$=[6,1]$). All experiments are implemented in PyTorch on NVIDIA 4090 GPUs.

\item {\bf Open access to data and code}
    \item[] Question: Does the paper provide open access to the data and code, with sufficient instructions to faithfully reproduce the main experimental results, as described in supplemental material?
    \item[] Answer: \answerNo{} \item[] Justification: The implementation is available at \url{https://github.com/Oryxo/MSAR}. The experiments use publicly available benchmark datasets, with descriptions and source references provided in Appendix~\ref{sec:dataset}. The repository includes a dependency list, data-format guidance, example training commands, configurable hyperparameters, and evaluation code. However, the current release does not provide complete data-preparation procedures and experiment-specific configurations for all reported results, so it does not yet satisfy the full reproducibility requirement of this question.
    \begin{itemize}
        \item The answer \answerNA{} means that paper does not include experiments requiring code.
        \item Please see the NeurIPS code and data submission guidelines (\url{https://neurips.cc/public/guides/CodeSubmissionPolicy}) for more details.
        \item While we encourage the release of code and data, we understand that this might not be possible, so \answerNo{} is an acceptable answer. Papers cannot be rejected simply for not including code, unless this is central to the contribution (e.g., for a new open-source benchmark).
        \item The instructions should contain the exact command and environment needed to run to reproduce the results. See the NeurIPS code and data submission guidelines (\url{https://neurips.cc/public/guides/CodeSubmissionPolicy}) for more details.
        \item The authors should provide instructions on data access and preparation, including how to access the raw data, preprocessed data, intermediate data, and generated data, etc.
        \item The authors should provide scripts to reproduce all experimental results for the new proposed method and baselines. If only a subset of experiments are reproducible, they should state which ones are omitted from the script and why.
        \item At submission time, to preserve anonymity, the authors should release anonymized versions (if applicable).
        \item Providing as much information as possible in supplemental material (appended to the paper) is recommended, but including URLs to data and code is permitted.
    \end{itemize}

\item {\bf Experimental setting/details}
    \item[] Question: Does the paper specify all the training and test details (e.g., data splits, hyperparameters, how they were chosen, type of optimizer) necessary to understand the results?
    \item[] Answer: \answerYes{} \item[] Justification: Training and evaluation details are described in the Experiments section and Appendix~\ref{appendix:exp_details}. This includes dataset splits following TSLib, the Adam optimizer with lr=$10^{-3}$ and early stopping (patience=3), unified \texttt{seq\_len}=336, and full hyperparameter search ranges.

\item {\bf Experiment statistical significance}
    \item[] Question: Does the paper report error bars suitably and correctly defined or other appropriate information about the statistical significance of the experiments?
    \item[] Answer: \answerNo{} \item[] Justification: The paper does not report error bars, confidence intervals, or statistical significance tests. The reported improvements are based on point estimates, and we do not systematically characterize variability across random seeds. Averaging over prediction lengths does not quantify seed-level uncertainty.

\item {\bf Experiments compute resources}
    \item[] Question: For each experiment, does the paper provide sufficient information on the computer resources (type of compute workers, memory, time of execution) needed to reproduce the experiments?
    \item[] Answer: \answerYes{} \item[] Justification: The appendix (Appendix~\ref{appendix:A3}) states that all experiments are conducted on two machines each equipped with four NVIDIA 4090 GPUs. Training efficiency (wall-clock time and GPU memory) is further analyzed in Figure~\ref{fig:parameters}.
    
\item {\bf Code of ethics}
    \item[] Question: Does the research conducted in the paper conform, in every respect, with the NeurIPS Code of Ethics \url{https://neurips.cc/public/EthicsGuidelines}?
    \item[] Answer: \answerYes{} \item[] Justification: This work proposes a time series forecasting framework evaluated on publicly available benchmark datasets. It does not involve human subjects, sensitive data, or any application with foreseeable ethical risks.

\item {\bf Broader impacts}
    \item[] Question: Does the paper discuss both potential positive societal impacts and negative societal impacts of the work performed?
    \item[] Answer: \answerNA{} \item[] Justification: This is a foundational research paper on time series forecasting methodology. There is no direct path to negative societal impacts from improving forecasting accuracy on standard benchmarks.
    
\item {\bf Safeguards}
    \item[] Question: Does the paper describe safeguards that have been put in place for responsible release of data or models that have a high risk for misuse (e.g., pre-trained language models, image generators, or scraped datasets)?
    \item[] Answer: \answerNA{} \item[] Justification: The paper presents a time series forecasting framework without releasing pre-trained models or scraped datasets. There is no high risk for misuse associated with this work.

\item {\bf Licenses for existing assets}
    \item[] Question: Are the creators or original owners of assets (e.g., code, data, models), used in the paper, properly credited and are the license and terms of use explicitly mentioned and properly respected?
    \item[] Answer: \answerYes{} \item[] Justification: All datasets and baseline models used in the paper are properly cited in the references. The datasets (ETT, ECL, Weather, Traffic, PEMS, etc.) are standard public benchmarks widely used in the time series forecasting literature.

\item {\bf New assets}
    \item[] Question: Are new assets introduced in the paper well documented and is the documentation provided alongside the assets?
    \item[] Answer: \answerNA{} \item[] Justification: The paper does not introduce new datasets or models released as assets. MSAR is a plug-in framework applied on top of existing models and evaluated on existing datasets.

\item {\bf Crowdsourcing and research with human subjects}
    \item[] Question: For crowdsourcing experiments and research with human subjects, does the paper include the full text of instructions given to participants and screenshots, if applicable, as well as details about compensation (if any)? 
    \item[] Answer: \answerNA{} \item[] Justification: The paper does not involve crowdsourcing or research with human subjects. All experiments are conducted on publicly available time series datasets.

\item {\bf Institutional review board (IRB) approvals or equivalent for research with human subjects}
    \item[] Question: Does the paper describe potential risks incurred by study participants, whether such risks were disclosed to the subjects, and whether Institutional Review Board (IRB) approvals (or an equivalent approval/review based on the requirements of your country or institution) were obtained?
    \item[] Answer: \answerNA{} \item[] Justification: The paper does not involve human subjects or crowdsourcing; hence IRB approval is not required.

\item {\bf Declaration of LLM usage}
    \item[] Question: Does the paper describe the usage of LLMs if it is an important, original, or non-standard component of the core methods in this research? Note that if the LLM is used only for writing, editing, or formatting purposes and does \emph{not} impact the core methodology, scientific rigor, or originality of the research, declaration is not required.
    \item[] Answer: \answerNA{} \item[] Justification: LLMs are not used as part of the core methodology of this research. MSAR is a scale-wise autoregressive framework for time series forecasting that does not rely on large language models.

\end{enumerate}

\end{document}